\documentclass[11pt]{article}
\usepackage[a4paper,top=2.75cm,bottom=2.75cm,left=2.5cm,right=2.5cm,marginparwidth=1.75cm]{geometry}
\usepackage[normalem]{ulem}
\PassOptionsToPackage{hyphens}{url}\usepackage[colorlinks = true,
            linkcolor = blue,
            urlcolor  = blue,
            citecolor = blue,
            anchorcolor = blue]{hyperref}     
\usepackage[none]{hyphenat}
\usepackage{breakcites}
\usepackage[utf8]{inputenc}
\usepackage{amsmath,amsthm}
\usepackage{amssymb}
\usepackage{mathtools}
\usepackage{soul}
\usepackage{scalerel}
\allowdisplaybreaks
\usepackage{tikz}
\usepackage{bm}
\usetikzlibrary{arrows}
\usetikzlibrary{arrows.meta}
\usepackage{dsfont}
\usepackage{graphicx, graphics}
\usepackage{wrapfig}
\usepackage{thmtools,thm-restate}
\usepackage{enumitem} %
\usepackage{booktabs}
\usepackage{multirow}
\usepackage[ruled,vlined]{algorithm2e}

\SetCommentSty{mycommfont}
\SetKwInput{KwInput}{Input}                %
\SetKwInput{KwOutput}{Output}  
\SetKwInput{KwInitialization}{Initialization}
\usepackage[noblocks]{authblk}

\usepackage[round]{natbib}

\usepackage[capitalize,noabbrev]{cleveref} 
\usepackage{autonum}

\newtheorem{remark}{Remark}
\newtheorem{definition}{Definition}
\newtheorem{assumption}{Assumption}
\newtheorem{theorem}{Theorem}
\newtheorem{corollary}{Corollary}
\newtheorem{lemma}{Lemma}

\newtheorem{proposition}{Proposition}

\def\tbf#1{\textbf{#1}}

\newcommand{\DO}{do}

\DeclareMathAlphabet{\mathbsf}{OT1}{cmss}{bx}{n}%
\DeclareMathAlphabet{\mathssf}{OT1}{cmss}{m}{sl}%

\def\wbar#1{\overline{#1}}

\newcommand{\rvx}{\mathssf{x}}	%
\newcommand{\rvv}{\mathssf{v}}	%
\newcommand{\rvu}{\mathssf{u}}	%
\newcommand{\rvz}{\mathssf{z}}	%
\newcommand{\rvt}{\mathssf{t}}	%
\newcommand{\rvy}{\mathssf{y}}	%
\newcommand{\rva}{\mathssf{a}}	%
\newcommand{\rvb}{\mathssf{b}}	%
\newcommand{\rvw}{\mathssf{w}}	%
\newcommand{\rvr}{\mathssf{r}}	%

\newcommand{\rvbs}{\mathbsf{s}}	%
\newcommand{\rvbr}{\mathbsf{r}}	%
\newcommand{\rvbx}{\mathbsf{x}}	%
\newcommand{\rvbz}{\mathbsf{z}}	%
\newcommand{\rvbv}{\mathbsf{v}}	%
\newcommand{\rvbu}{\mathbsf{u}}	%
\newcommand{\rvbw}{\mathbsf{w}}	%
\newcommand{\rvbb}{\mathbsf{b}}	%

\newcommand{\rvzi}{\rvz^{(i)}}
\newcommand{\rvbzo}{\rvbz^{(o)}}	
\newcommand{\rvbzi}{\rvbz^{(i)}}

\newcommand{\svbz}{\boldsymbol{z}} %
\newcommand{\svbb}{\boldsymbol{b}} %
\newcommand{\svbs}{\boldsymbol{s}} %
\newcommand{\svbr}{\boldsymbol{r}} %

\newcommand{\Indicator}{\mathds{1}}

\newcommand{\svbzo}{\svbz^{(o)}}		
\newcommand{\svbzi}{\svbz^{(i)}}
\newcommand{\svbu}{\boldsymbol{u}}

\newcommand{\urvt}{\underline{\rvt}}

\newcommand{\brvbv}{\wbar{\rvbv}}	
\newcommand{\urvbv}{\underline{\rvbv}}

\newcommand{\brvbb}{\wbar{\rvbb}}	
\newcommand{\urvbb}{\underline{\rvbb}}

\newcommand{\cZ}{\mathcal{Z}}
\newcommand{\cW}{\mathcal{W}}
\newcommand{\cG}{\mathcal{G}}
\newcommand{\cP}{\mathcal{P}}
\newcommand{\cM}{\mathcal{M}}
\newcommand{\cU}{\mathcal{U}}
\newcommand{\cV}{\mathcal{V}}

\newcommand{\Probability}{\mathbb{P}}
\newcommand{\Expectation}{\mathbb{E}}

\newcommand{\dsep}{\hspace{1mm}{\perp \!\!\! \perp}_d~}
\newcommand{\indep}{\perp_p}

\newcommand{\doublearrow}{{~\dashleftarrow \hspace{-5.5mm} \dashrightarrow~}}
\newcommand{\myrightarrow}{\xrightarrow{\hspace*{2.5mm}}}
\newcommand{\myleftarrow}{\xleftarrow{\hspace*{2.5mm}}}

\newcommand{\sequal}[1]{\stackrel{#1}{=}}

\newcommand{\normalbrackets}[1]{[ #1 ]}

\newcommand{\Bigbrackets}[1]{\Big[ #1 \Big]}

\newcommand{\bigparenth}[1]{\big( #1 \big)}
\newcommand{\Bigparenth}[1]{\Big( #1 \Big)}

\newcommand{\normalbraces}[1]{\{ #1  \}}

\newcommand{\normalabs}[1]{| #1  |}

\def\defeq{\triangleq} %
\newcommand{\defn}{\defeq}
\newcommand{\qtext}[1]{\quad\text{#1}\quad} 
\newcommand{\stext}[1]{\ \text{#1}\ }

\newcommand{\doot}[1][t]{do(\rvt = #1)}
\newcommand{\doob}[1][\svbb]{do(\rvbb = #1)}

\newcommand{\pao}{\pi^{(o)}} %
\newcommand{\pau}{\pi^{(u)}} %

\newcommand{\cGtoy}{\mathcal{G}^{toy}}

\newtheorem{condition}{Condition}
\newtheorem{problem}{Problem}
\newtheorem{fact}{Fact}

\newtheorem{myproposition}{Proposition}

\newtheorem{mycorollary}{Corollary}

\newtheorem{mylemma}{Lemma}

\newtheorem{mydefinition}{Definition}

 \newtheorem{myproblem}{Problem}

\newtheorem{myremark}{Remark}

\newtheorem{myassumption}{Assumption}

\newtheorem{mycondition}{Condition}

\newtheorem{myfact}{Fact}

 \crefname{appendix}{Appendix}{Appendices}
\crefname{equation}{}{}
\crefname{lemma}{Lemma}{Lemmas}
\crefname{theorem}{Theorem}{Theorems}
\crefname{Corollary}{Corollary}{Corollaries}
\crefname{algorithm}{Algorithm}{Algorithms}

\crefname{section}{Section}{Sections}
\crefname{table}{Table}{Tables}
\crefname{remark}{Remark}{Remarks}
\crefname{definition}{Definition}{Definitions}

\crefname{Proposition}{Proposition}{Propositions}
\crefname{myproblem}{Problem}{Problems}
\crefname{myremark}{Remark}{Remarks}
\crefname{mylemma}{Lemma}{Lemmas}
\crefname{mydefinition}{Definition}{Definitions}
\crefname{myproposition}{Proposition}{Propositions}
\crefname{mycorollary}{Corollary}{Corollaries}
\crefname{myassumption}{Assumption}{Assumptions}
\crefname{mycondition}{Condition}{Conditions}
\crefname{myfact}{Fact}{Facts}
\crefname{figure}{Figure}{Figures}
\crefname{enumi}{}{}
\crefname{name}{}{} %

\graphicspath{{../figures/},{figures/}}

\title{Front-door Adjustment Beyond Markov Equivalence with Limited Graph Knowledge}
\author[1]{Abhin Shah}
\author[2]{Karthikeyan Shanmugam}
\author[3]{Murat Kocaoglu}

\affil[1]{Massachusetts Institute of Technology}
\affil[2]{Google Research India}
\affil[3]{Purdue University}
\date{}
\begin{document}
\sloppy
\maketitle
\begin{abstract}
Causal effect estimation from data typically requires assumptions about the cause-effect relations either explicitly in the form of a causal graph structure within the Pearlian framework, or implicitly in terms of (conditional) independence statements between counterfactual variables within the potential outcomes framework. When the treatment variable and the outcome variable are confounded, front-door adjustment is an important special case where, given the graph, causal effect of the treatment on the target can be estimated using \textit{post-treatment} variables. However, the exact formula for front-door adjustment depends on the structure of the graph, which is difficult to learn in practice. In this work, we provide testable conditional independence statements to compute the causal effect using front-door-like adjustment without knowing the graph under limited structural side information. We show that our method is applicable in scenarios where knowing the Markov equivalence class is not sufficient for causal effect estimation. We demonstrate the effectiveness of our method on a class of random graphs as well as real causal fairness benchmarks. 
\end{abstract}
\section{Introduction}
\label{sec_intro}
Causal effect estimation is at the center of numerous scientific, societal, and medical questions~\citep{nabi2019learning,castro2020causality}. The $\DO(\cdot)$ operator of Pearl represents the effect of an experiment on a causal system. For example, the probability distribution of a target variable $\rvy$ after setting a treatment $\rvt$ to $t$ is represented by $\Probability(\rvy|\DO(\rvt=t))$ and is known as an interventional distribution. Learning this distribution for any realization $\rvt=t$\footnote{Depending on the context, causal effect estimation sometimes refers to the estimating the difference of assigning $\rvt=1$ vs. $\rvt=0$ on the target variable $\rvy$, e.g., $\mathbb{E}[\rvy|\DO(\rvt=1)]-\mathbb{E}[\rvy|\DO(\rvt=0)]$. This quantity is computable if we can identify $\Probability(\rvy|\DO(\rvt=t))$ for $t = \{0,1\}$.} is what causal effect estimation entails. This distribution is different from the conditional distribution $\Probability(\rvy|\rvt=t)$ as there may be unobserved confounders between treatment and outcome that cannot be controlled for. 

\begin{figure}[t]
    \centering
    \begin{tabular}{c}
	\begin{tikzpicture}[scale=0.75, every node/.style={transform shape}, > = latex, shorten > = 1pt, shorten < = 1pt]
	\node[shape=circle,draw=black](y) at (6,-1.5) {\LARGE$\rvy$};
	\node[shape=circle,draw=black](t) at (4,-1.5) {\LARGE$\rvt$};
	\node[shape=circle,draw=black](z) at (2,-1.5) {\LARGE$\rvz$};
	\path[style=thick][<->, bend left](z) [dashed] edge (y);
	\path[style=thick][->](t) edge (y);
	\path[style=thick][->](z) edge (t);
	\end{tikzpicture} \qquad \qquad \qquad
    \begin{tikzpicture}[scale=0.75, every node/.style={transform shape}, > = latex, shorten > = 1pt, shorten < = 1pt]
	\node[shape=circle,draw=black](y) at (6,-1.5) {\LARGE$\rvy$};
	\node[shape=circle,draw=black](z) at (2,-1.5) {\LARGE$\rvt$};
	\node[shape=circle,draw=black](t) at (4,-1.5) {\LARGE$\rvz$};
	\path[style=thick][<->, bend left](z) [dashed] edge (y);
	\path[style=thick][->](t) edge (y);
	\path[style=thick][->](z) edge (t);
	\end{tikzpicture} \\
    \end{tabular}
    \caption{Representative graphs for back-door adjustment {(left)} and front-door adjustment {(right)}.} 
    \label{fig_graphical_models}
\end{figure}
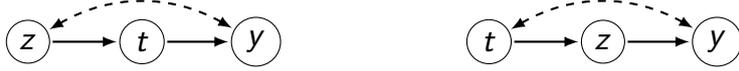

A causal graph, often depicted as a directed acyclic graph, captures the cause-and-effect relationships between variables and explains the causal system under consideration. A semi-Markovian causal model represents a causal model that includes unobserved variables influencing multiple observed variables  \citep{verma1990causal, acharya2018learning}. In a semi-Markovian graph, directed edges between observed variables represent causal relationships, while bi-directed edges between observed variables represent unobserved common confounding (see Figure \ref{fig_graphical_models}). Given any semi-Markovian graph, complete identification algorithms for causal effect estimation are known. For example, if $\Probability(\rvy|do(\rvt=t))$ is uniquely determined by the observational distribution and the causal graph, the algorithm by \citet{shpitser2006identification} utilizes the graph to derive an \emph{estimand}, i.e., the functional form mapping the observational distribution to the interventional distribution.

Certain special cases of estimands have found widespread use across several domains. One such special case is the \emph{back-door adjustment} \citep{Pearl1993} shown in Figure \ref{fig_graphical_models}{(left)}. The back-door adjustment utilizes the pre-treatment variable $\rvbz$ (that blocks back-door paths) to control for unobserved confounder as follows: 
\begin{align}
    \Probability(\rvy|\DO(\rvt = t)) = \sum_{\svbz} \Probability(\rvy|\rvt = t,\rvbz = \svbz)\Probability(\rvbz = \svbz),
\end{align} 
where the do-calculus rules of \cite{pearl1995causal} are used to convert interventional distributions into observational distributions by leveraging the graph structure. However, the back-door adjustment is often inapplicable, e.g., in the presence of an unobserved confounder between $\rvt$ and $\rvy$. Surprisingly, in such scenarios, it is sometimes possible to find the causal effect using the \emph{front-door adjustment} \citep{pearl1995causal} shown in Figure \ref{fig_graphical_models}{(right)}. Utilizing the front-door variable $\rvbz$, the front-door adjustment estimates the causal effect from observational distributions using the following formula (which is also obtained through the do-calculus rules and the graph structure):
\begin{align}
    \Probability(\rvy|\DO(\rvt =t)) \!=\!\sum_{\svbz} \Bigparenth{\sum_{t'} \Probability(\rvy|\rvt = t',\rvbz = \svbz)\Probability(\rvt=t')} \Probability(\rvbz = \svbz|\rvt=t)
\end{align}
Recently, front-door adjustment has gained popularity in analyzing real-world data \citep{glynn2017front,bellemare2019paper,hunermund2019causal} due to its ability to utilize post-treatment variables to estimate effects even in the presence of confounding between $\rvt$ and $\rvy$. However, in general, front-door adjustment also relies on knowing the causal graph, which may not always be feasible, especially in domains with many variables.

{An alternative approach uses observational data to infer a Markov equivalence class, which is a collection of causal graphs that encode the same conditional independence relations \citep{spirtes2000causation}. A line of work \citep{perkovic2018complete,jaber2019causal} provide identification algorithms for causal effect estimation from partial ancestral graphs (PAGs) \citep{zhang2008completeness}, a prominent representation of the Markov equivalence class, whenever every causal graph in the collection shares the same causal effect estimand. However, learning PAGs from data is challenging in practice due to the sequential nature of their learning algorithms, which can propagate errors between tests \citep{strobl2019estimating}. Further, to the best of our knowledge, there is no existing algorithm that can incorporate side information, such as known post-treatment variables, into PAG structure learning.}

In this work, we ask the following question: \textit{Can the causal effect be estimated with a testable criteria on observational data by utilizing some structural side information without knowing the graph?}

Recent research has developed such testable criteria to enable back-door adjustment without knowing the full causal graph \citep{entner2013data, cheng2020towards, gultchin2020differentiable, shah2022finding}. These approaches leverage structural side information, such as a known and observed parent of the treatment variable $\rvt$. However, no such results have been established for enabling front-door adjustment. We address this gap by focusing on the case of unobserved confounding between $\rvt$ and $\rvy$, where back-door adjustment is inapplicable. Traditionally, this scenario has been addressed by leveraging the presence of an instrumental variable \citep{mogstad2018identification} or performing sensitivity analysis \citep{veitch2020sense}, both of which provide only bounds in the non-parametric case. In contrast, we achieve identifiability by utilizing structural side information.\\

\noindent \textbf{Contributions.}  
We propose a method for estimating causal effects without requiring the knowledge of causal graph in the presence of unobserved confounding between treatment and outcome. Our approach utilizes front-door-like adjustments based on post-treatment variables and relies on conditional independence statements that can be directly tested from observational data. We require one structural side information which can be obtained from an expert and is less demanding than specifying the entire causal graph. We illustrate that our framework provides identifiability in random ensembles where existing PAG-based methods are not applicable. Further, we illustrate the practical application of our approach to causal fairness analysis by estimating the total effect of a sensitive attribute on an outcome variable using the German credit data with fewer structural assumptions.

\subsection{Related Work}
\label{sec_related_work}
\textbf{Effect estimation from causal graphs/Markov equivalence Class:} The problem of estimating interventional distributions with the knowledge of the semi-Markovian model has been studied extensively in the literature, with important contributions such as \citet{tian2002general} and \citet{shpitser2006identification}. \citet{perkovic2018complete} presented a complete and sound algorithm for identifying valid adjustments from PAGs. Going beyond valid adjustments, \citet{jaber2019causal} proposed a complete and sound algorithm for identifying causal effect from PAGs. However, our method can recover the causal effect in scenarios where these algorithms are inapplicable.\\

\noindent \textbf{Effect estimation via front-door adjustment with causal graph:} 
Several recent works have contributed to a better understanding of the statistical properties of front-door estimation \citep{kuroki2000selection, kuroki2012selection, glynn2018front, gupta2021estimating}, proposed robust generalizations \citep{hunermund2019causal, fulcher2020robust}, and developed procedures to enumerate all possible front-door adjustment sets \citep{jeong2022finding, wienobst2022finding}. However, all of these require knowing the underlying causal graph. In contrast, \citet{bhattacharya2022testability} verified the front-door criterion without knowing the causal graph using Verma constraint-based methodology. However, this approach is only applicable to a small set of graphs. Our proposed approach relies on conditional independence and is applicable to a broad class of graphs.
\section{Preliminaries and Problem Formulation}
\label{sec_prob_formulation}
\noindent{\bf Notations.} 
For a sequence of realizations %
$r_1, \cdots , r_n$, we define $\svbr \defn \{r_1, \cdots, r_n\}$. For a sequence of random variables $\rvr_1, \cdots , \rvr_n$, we define $\rvbr \defn \{\rvr_1, \cdots, \rvr_n\}$. Let $\Indicator$ denote the indicator function.\\

\noindent{\bf Semi-Markovian Model and Effect Estimation.}
We consider a causal effect estimation task where $\rvbx$ represents the set of observed features, $\rvt$ represents the observed treatment variable, and $\rvy$ represents the observed outcome variable. We denote the set of all observed variables jointly by $\cV \defn \{\rvbx,\rvt,\rvy\}$. Let $\cU$ denote the set of unobserved features that could be correlated with the observed variables.\\

\noindent We assume $\cW \defn \cV \cup \cU$ follows a semi-Markovian causal model \citep{tian2002general} as below. 
\begin{definition} 
A semi-Markovian causal model (SMCM) $\cM$ is specified as follows: 
\begin{enumerate}[leftmargin=*,topsep=-3pt,itemsep=-4pt]
    \item  $\cG$ is a directed acyclic graph (DAG) over the set of vertices $\cW$ such that each element of the set $\cU$ has no parents.
    \item $\forall \rvv \in \cV$, let $\pao(\rvv) \subseteq \cV$ and $\pau(\rvv) \subseteq \cU$ denote the set of parent of $\rvv$ in $\cV$ and $\cU$, respectively.
    \item  $\Probability(\rvbu)$ is the unobserved joint distribution over the unobserved features.
    \item The observational distribution is given by $\Probability(\rvbv) = \Expectation_{\svbu} \Bigbrackets{\prod \limits_{\rvv \in \cV} \Probability(\rvv|\pi^{(o)}(\rvv),\pi^{(u)}(\rvv))}$.
    \item The interventional distribution when the variables $\rvbr \subset \cV$ are set to a fixed value $\svbr$ is given by
       \begin{align}\label{eq:do-dist}
         \Probability(\rvbv|do(\rvbr = \svbr)) = \Indicator_{\rvbr=\svbr} \cdot \Expectation_{\svbu} \Bigbrackets{\prod \limits_{\rvv \in \cV \setminus \rvbr} \Probability(\rvv|\pi^{(o)}(\rvv),\pi^{(u)}(\rvv))}. 
       \end{align}
    \item For any $\rvv_1,\rvv_2 \in \cV$, if $\pi^{(u)}(\rvv_1) \cap \pi^{(u)}(\rvv_2) \neq \emptyset$, then $\rvv_1$ and $\rvv_2$ have a bi-directed edge in $\cG$.     
\end{enumerate}
\end{definition}

\noindent In this work, we are interested in the causal effect of $\rvt$ on $\rvy$, i.e., $\Probability(\rvy|\DO(\rvt=t))$. We define this formally by marginalizing all variables except $\rvy$ in the interventional distribution in \eqref{eq:do-dist}.
\begin{definition}
\label{def:t-effect}
The causal effect of $\rvt$ (when forced to a value $t$) on $\rvy$ is given by: 
 \begin{align}
     \Probability(\rvy|do(\rvt=t))  \!=\!\! \sum \limits_{\rvbv \setminus\{\rvy\}}\Probability \bigparenth{\rvbv \setminus\{\rvy\}, \rvy | do(\rvt=t)}.
 \end{align}
\end{definition}
\noindent Next, we define the notion of average treatment effect for a binary treatment $\rvt$.
 \begin{definition}\label{definition_average_treatment_effect}
 The average treatment effect (ATE) of a binary treatment $\rvt$ on outcome $\rvy$ is given by ATE = $\Expectation[\rvy | do(\rvt = 1)] - \Expectation[\rvy | do(\rvt = 0)]$.
 \end{definition}

\noindent Next, we define when the causal effect (Definition \ref{def:t-effect}) is said to be identifiable from the observational distribution and the causal graph.  
\begin{definition}\label{def:ID}
({Causal effect identifiability})
Given an observational distribution $\Probability(\rvbv)$ and a causal graph $\cG$, the causal effect $\Probability(\rvy | do(\rvt =t))$ is identifiable if it is identical for every semi-Markovian Causal model with $(a)$ same graph $\cG$ and $(b)$ same observational distribution $\Probability(\rvbv)$.
\end{definition}
\noindent In a causal graph $\cG$, a path is an ordered sequence of distinct nodes where each node is connected to the next in the sequence by an edge. A path starting at node $\rvw_1$ and ending at node $\rvw_2$ in $\cG$ is \textit{blocked} by a set $\rvbw \subset \cW \setminus \{\rvw_1, \rvw_2\}$ if there exists $\rvw \in \rvbw$ such that (a) $\rvw$ is not a collider or (b) $\rvw$ is a collider and neither $\rvw$ nor any of it’s descendant is in $\rvbw$. Further, $\rvw_1$ and $\rvw_2$ are said to be \textit{d-separated} by $\rvbw$ in $\cG$ if $\rvbw$ blocks every path between $\rvw_1$ and $\rvw_2$ in $\cG$. Let $\rvw_1 \dsep \rvw_2 |\rvbw$ denote that $\rvw_1$ and $\rvw_2$ are d-separated by $\rvbw$ in $\cG$. Similarly, let $\rvw_1 \indep \rvw_2 |\rvbw$ denote that $\rvw_1$ and $\rvw_2$ are conditionally independent given $\rvbw$. We assume causal faithfulness, i.e., any conditional independence $\rvw_1 \indep \rvw_2 |\rvbw$ implies a d-separation relation $\rvw_1 \dsep \rvw_2 |\rvbw$ in the causal graph $\cG$. 

\subsection{Adjustment using pre-treatment variables}
It is common in causal effect estimation  to consider pre-treatment variables, i.e., variables that occur before the treatment in the causal ordering, and identify sets of variables that are \textit{valid adjustments}. Specifically, a set $\rvbz \subset \cV$ forms a valid adjustment if the causal effect can be written as $\Probability(\rvy|do(\rvt =t)) = \sum_{\svbz} \Probability(\rvy|\rvt = t,\rvbz = \svbz)\Probability(\rvbz = \svbz)$. In other words, a valid adjustment $\rvbz$ averages an estimate of $\rvy$ regressed on $\rvt$ and $\rvbz$ with respect to the marginal distribution of $\rvbz$. A popular criterion to find valid adjustments is to find a set $\rvbz \subset \cV$ that satisfies the \textit{back-door criterion} \citep{pearl2009causality}. Formally, a set $\rvbz$ satisfies the back-door criterion if (a) it blocks all back-door paths, i.e., paths between $\rvt$ and $\rvy$ that have an arrow pointing at $\rvt$ and (b) no element of $\rvbz$ is a descendant of $\rvt$. While, in general, back-door sets can be found with the knowledge of the causal graph, recent works (see the survey \citet{cheng2022data}) have proposed testable criteria for identifying back-door sets with some causal side information, without requiring the entire graph.

\subsection{Adjustment using post-treatment variables}
While back-door adjustment is widely used, there are scenarios where no back-door set exists, e.g., when there is an unobserved confounder between $\rvt$ and $\rvy$. If no back-door set can be found from the pre-treatment variables, Pearlian theory can be used to identify post-treatment variables, i.e., the variables that occur after the treatment in the causal ordering, to obtain a \textit{front-door adjustment}.
\begin{definition}[\textit{Front-door criterion}]
\label{def_fd}
A set $\rvbz \subset \cV$ satisfies the front-door criterion with respect to $\rvt$ and $\rvy$ if (a) $\rvbz$ intercepts all directed paths from $\rvt$ to $\rvy$ (b) all back-door paths between $\rvt$ and $\rvbz$ are blocked, and (c) all back-door paths between $\rvbz$ and $\rvy$ are blocked by $\rvt$.
\end{definition}
\noindent If a set $\rvbz$ satisfies the front-door criterion, then the causal effect can be written as 
\begin{align}\label{eqn:front-door}
    \Probability(\rvy|\DO(\rvt =t)) \!=\!\sum_{\svbz} \Bigparenth{\sum_{t'} \Probability(\rvy|\rvt = t',\rvbz = \svbz)\Probability(\rvt=t')} \Probability(\rvbz = \svbz|\rvt=t).
\end{align}
Intuitively, front-door adjustment estimates the causal effect of $\rvt$ on $\rvy$ as a composition of two effects: $(a)$ the effect of $\rvt$ on $\rvbz$ and $(b)$ the effect of $\rvbz$ on $\rvy$. However, one still needs the knowledge of the causal graph $\cG$ to find a set satisfying the front-door criterion.

Inspired by the progress in finding back-door sets without knowing the entire causal graph, we ask: \textit{Can testable conditions be derived to identify front-door-like sets using only partial structural information about post-treatment variables?} To that end, we consider the following side information.
\begin{assumption}\label{assumption_descendant}
The outcome $\rvy$ is a descendant of the treatment  $\rvt$.
\end{assumption}
\begin{assumption}\label{assumption_confounded} 
There is an unobserved confounder between the outcome $\rvy$ and the treatment  $\rvt$.
\end{assumption}
\begin{assumption}\label{assumption_known_children}
$\rvbb$, the set of all children of  the treatment $\rvt$, is observed and known.
\end{assumption}

Assumption \ref{assumption_descendant} is a fundamental assumption in most causal inference works, as it forms the basis for estimating non-trivial causal effects. Without it, the causal effect would be zero. Assumption \ref{assumption_confounded} rules out the existence of sets that satisfy the back-door criteria, necessitating a different way of estimating the causal effect. Assumption \ref{assumption_known_children} captures our side information by requiring every children of the treatment to be known and observed. To contrast, the side information in data-driven works on back-door adjustment requires a parent of the treatment to be known and observed \citep{shah2022finding}.

\begin{figure}[t]
  \centering
	\begin{tikzpicture}[scale=0.7, every node/.style={transform shape}, > = latex, shorten > = 1pt, shorten < = 1pt]
	\node[shape=circle,draw=black](u1) at (0,2) {\LARGE$\rvu_1$};
	\node[shape=circle,draw=black](u2) at (0,0) {\LARGE$\rvu_2$};
	\node[shape=circle,draw=black](u3) at (2,0) {\LARGE$\rvu_3$};
	\node[shape=circle,draw=black](u4) at (5,1) {\LARGE$\rvu_4$};
	\node[shape=circle,draw=black](u5) at (2,-3.5) {\LARGE$\rvu_5$};
	\node[shape=circle,draw=black](x1) at (-2,2) {\LARGE$\rvx_1$};
	\node[shape=circle,draw=black](x2) at (2,2) {\LARGE$\rvx_2$};
	\node[shape=circle,draw=black](y) at (6,-2) {\LARGE$\rvy$};
	\node[shape=circle,draw=black](ztilde) at (4,-2) {\Large$\rvbzo$};
	\node[shape=circle,draw=black](zhat) at (4,0) {\Large$\rvbzi$};
	\node[shape=circle,draw=black](b) at (2,-2) {\Large$\rvbb$};
	\node[shape=circle,draw=black](t) at (-2,-2) {\LARGE$\rvt$};
	\path[style=thick][->](x1) edge (t);
	\path[style=thick][->](t) edge (b);		
	\path[style=thick][->](b) edge (ztilde);
	\path[style=thick][->](ztilde) edge (y);
	\path[style=thick][<-](b) edge (zhat);
	\path[style=thick][->](zhat) edge (y);
	\path[style=thick][->](u3) [dashed] edge (x2);
	\path[style=thick][->](u3) [dashed] edge (b);
	\path[style=thick][->](u2) [dashed] edge (t);
	\path[style=thick][->](u2) [dashed] edge (x2);		
	\path[style=thick][->](u1) [dashed] edge (x1);
	\path[style=thick][->](u1) [dashed] edge (x2);
	\path[style=thick][->, bend left=15](u5) [dashed] edge (t);
	\path[style=thick][->, bend right=15](u5) [dashed] edge (y);
	\path[style=thick][->, bend right=20](u4) [dashed] edge (x2);
	\path[style=thick][->, bend left=20](u4) [dashed] edge (y);
	\end{tikzpicture}
  \caption{The graph $\cGtoy$ satisfying Assumptions \ref{assumption_descendant} to \ref{assumption_known_children} where $\rvu_i$ are unobserved.}
  \label{fig_toy_example}
\end{figure}
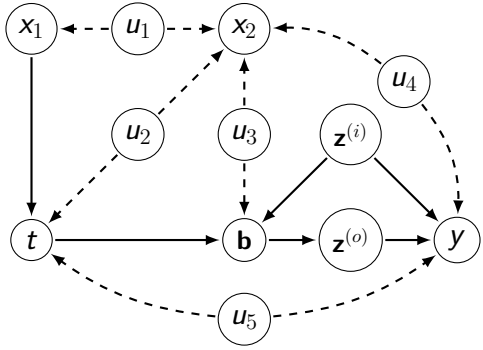
Our assumptions imply that $\rvbb$ intercepts all the directed paths from $\rvt$ to $\rvy$. Given this, it is natural to ask whether $\rvbb$ satisfies the front-door criterion (Definition \ref{def_fd}). We note that, in general, this is not true. We illustrate this via Figure \ref{fig_toy_example} where we provide a causal graph $\cGtoy$ satisfying our assumptions. However, $\rvbb$ is not a valid front-door set in $\cGtoy$ as the back-door path between $\rvbb$ and $\rvy$ via $\rvbzi$ is not blocked by $\rvt$. Therefore, estimating the causal effect by assuming $\rvbb$ is a front-door set might not always give an unbiased estimate. In the next section, we leverage the given side information and provide testable conditions to identify front-door-like sets.
\section{Front-door Adjustment Beyond Markov Equivalence}
In this section, we provide our main results, an algorithm for ATE estimation, and discuss the relationship to PAG-based methods. Our main results use observational criteria for causal effect estimation under Assumptions \ref{assumption_descendant} to \ref{assumption_known_children} using post treatment variables.
\subsection{Causal effect estimation using post-treatment variables}
\label{subsubsec_causal_effect_Estimation_post_treatment}
First, we state a conditional independence statement implying causal identifiability. Then, we provide additional conditional independence statements resulting in a unique formula for effect estimation.

\begin{figure}
  \centering
  \vspace{-4.5mm}
 \begin{tabular}{cc}
	\begin{tikzpicture}[scale=0.6, every node/.style={transform shape}, > = latex, shorten > = 1pt, shorten < = 1pt]
	\node[shape=circle,draw=black](y) at (6,-1.5) {\LARGE$\rvy$};
	\node[shape=circle,draw=black](ztilde) at (4,-1.5) {\Large$\rvz_2$};
	\node[shape=circle,draw=black](zhat) at (1.0,0.5) {\Large$\rvz_1$};
	\node[shape=circle,draw=black](b) at (2,-1.5) {\Large$\rvbb$};
	\node[shape=circle,draw=black](t) at (0,-1.5) {\LARGE$\rvt$};
	\path[style=thick][<->, bend right](t) [dashed] edge (y);
	\path[style=thick][->](t) edge (b);
	\path[style=thick][->](b) edge (ztilde);
	\path[style=thick][->](ztilde) edge (y);
	\path[style=thick][<-](b) edge (zhat);
	\path[style=thick][->](zhat) edge (t);
	\end{tikzpicture} \qquad \qquad \qquad \begin{tikzpicture}[scale=0.6, every node/.style={transform shape}, > = latex, shorten > = 1pt, shorten < = 1pt]
	\node[shape=circle,draw=black](y) at (6,-1.5) {\LARGE$\rvy$};
	\node[shape=circle,draw=black](ztilde) at (4,-1.5) {\Large$\rvz_2$};
	\node[shape=circle,draw=black](b) at (2,-1.5) {\Large$\rvbb$};
	\node[shape=circle,draw=black](t) at (0,-1.5) {\LARGE$\rvt$};
	\path[style=thick][<->, bend right](t) [dashed] edge (y);
	\path[style=thick][->](t) edge (b);
	\path[style=thick][->](b) edge (ztilde);
	\path[style=thick][->](ztilde) edge (y);
	\end{tikzpicture} 
  \end{tabular}
  \vspace{-3mm}
  \caption{SMCMs on {(left)} \& {(right)} satisfy $\rvbb \dsep \rvy | \rvt, \rvbz$ but have different causal effect estimation formulae.}
 \label{fig:smcm_fail_condition}
\end{figure}
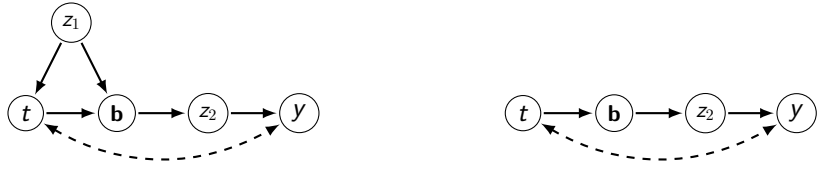

Causal identifiability (Definition \ref{def:ID}) implies that the causal effect is  uniquely determined given an observational distribution $\Probability(\cV)$ and the corresponding causal graph $\cG$. We now show that satisfying a conditional independence statement (which can be tested solely from observational data, without requiring the graph $\cG$) guarantees identifiability. We provide a proof in Appendix \ref{sec_proof_thm_tian}.
\newcommand{\id}{Causal Identifiability}
\begin{theorem}[\tbf{\id}]\label{thm_tian}
    Suppose Assumptions \ref{assumption_descendant} to \ref{assumption_known_children} hold. If there exists a set $\rvbz  \subseteq \cV \setminus \{\rvt, \rvbb, \rvy\}$ such that $\rvbb \dsep \rvy | \rvt, \rvbz$, then the causal effect of $\rvt$ on $\rvy$ is identifiable from observational data without the knowledge of the underlying causal graph $\cG$.
\end{theorem}
While the above result leads to identifiability, it does not provide a formula to compute the causal effect. In fact, the conditional independence $\rvbb \dsep \rvy | \rvt, \rvbz$ alone is \textit{insufficient} to establish a unique formula, and different causal graphs lead to different formula. To illustrate this, we provide two SMCMs where Assumptions \ref{assumption_descendant} to \ref{assumption_known_children} and $\rvbb \dsep \rvy | \rvt, \rvbz$ hold, i.e., causal effect is identifiable from observational data via Theorem \ref{thm_tian}, but the formula is different. First, consider the SMCM in Figure \ref{fig:smcm_fail_condition}{(left)} with $\rvbz = (\rvz_1, \rvz_2)$ where causal effect is given by following formula (derived in Appendix \ref{sec_proof_thm_tian}): 
\begin{align}\label{eq_diff_formula}
    \Probability(\rvy | \doot) = \sum_{z_1, z_2, \svbb} \Bigparenth{\sum_{t'} \Probability(\rvy |  z_1, z_2, t') \Probability( t' | z_1)} \times \Probability(z_2 |\svbb)  \Probability(\svbb |  t,  z_1) \Probability(z_1).
\end{align}
Next, consider the SMCM in Figure \ref{fig:smcm_fail_condition}{(right)} with $\rvbz = \rvz_2$ where the causal effect is given by the front-door adjustment formula in \eqref{eqn:front-door} as $\rvbz$ satisfies the front-door criterion. It remains to explicitly show that the formula in \eqref{eq_diff_formula} is different from \eqref{eqn:front-door}. To this end, we create a synthetic structural equation model (SEM) respecting the graph in Figure \ref{fig:smcm_fail_condition}{(left)} and show that the formula in \eqref{eqn:front-door} gives a non-zero ATE error. In our SEM, the unobserved variable has a uniform distribution over $[1,2]$. Each observed variable except $\rvt$ is a sum of $(i)$ a linear combination of its parents with  coefficients drawn from uniform distribution over $[1,2]$ and $(ii)$ a zero-mean Gaussian noise. The treatment variable is binarized by applying a standard logistic model to a linear combination of its parents with coefficients drawn as before. The ATE error averaged over 50 runs with 50000 samples in each run is $0.3842 \pm 0.0207$. See more experimental details in Appendix \ref{appendix_experiments}.

Next, we provide two additional conditional independence statements that imply a unique formula for causal effect estimation. Our result is a \textit{generalized front-door} with a formula identical to \eqref{eqn:front-door} as if $\rvbz$ were a traditional front-door set. We also offer an alternative formula by utilizing a specific partition of $\rvbz$ obtained from the conditional independence statements. We provide a proof in Appendix \ref{sec_proof_thm_main}.
\newcommand{\gfd}{A generalized front-door condition}
\begin{theorem}[\tbf{\gfd}]\label{thm_main}
Suppose \cref{assumption_descendant,assumption_confounded,assumption_known_children} hold. Let $\rvbz \subseteq \cV \setminus \{\rvt, \rvbb, \rvy\}$ be a set satisfying 
\begin{align}
    \rvbb \dsep \rvy | \rvt, \rvbz, \label{eq_ci}
\end{align} 
such that $\rvbz$ can be decomposed into $\rvbzo \subseteq \rvbz$ and $\rvbzi = \rvbz \setminus \rvbzo$ with
\begin{align}
\stext{(i)}\rvbzi ~ \dsep \rvt  \qtext{and} \stext{(ii)} \rvbzo ~ \dsep \rvt | \rvbb, \rvbzi. \label{eq_condition}
\end{align}
Then, $\rvbz$ and $\rvbs \defn (\rvbb, \rvbzi)$ are generalized front-doors, and the causal effect of $\rvt$ on $\rvy$ can be obtained using any of the following equivalent formulae:
\begin{align}
    \Probability(\rvy | \doot) & = \sum_{\svbz} \Bigparenth{\sum_{t'} \Probability(\rvy | \svbz, t') \Probability(t')} \Probability(\svbz | t). \label{eq_thm_main_1}\\
    \Probability(\rvy| \doot) & = \sum_{\svbs} \Bigparenth{\sum_{t'} \Probability(\rvy |  \svbs, t') \Probability( t')} \Probability( \svbs |t). \label{eq_thm_main_2}
\end{align}
\end{theorem}
\subsection{Algorithm for ATE estimation}
\begin{algorithm}[h]
\KwInput{ $n_r, \rvt, \rvy, \rvbb, \cZ, p_{v}$}
\KwOutput{$\text{ATE}_z, \text{ATE}_s$}
\KwInitialization{$\text{ATE}_z = 0, \text{ATE}_s = 0, c_1 = 0$ }
\For(\tcp*[h]{Use a different train-test split in each run}) {$r = 1,\cdots,n_r$} 
{ $\text{ATE}^{r}_z = 0$, $\text{ATE}^{r}_s = 0$, $c_2 = 0$; \\
  \For {$\rvbz \in \cZ$}
    {
    \If (\tcp*[h]{where $CI$ stands for conditional independence}) {$CI(\rvbb \indep \rvy | \rvbz, t) > p_{v}$}
        {
        \For {$\rvbzo \subseteq \rvbz$}
            {
            $\rvbzi = \rvbz \setminus \rvbzo$;\\
            \If {$\min\{CI(\rvbzi \!\indep\! \rvt),CI(\rvbzo \!\indep\! \rvt | \rvbb, \rvbzi)\} > p_{v})$} 
                {
                $c_2 = c_2 + 1$, $\rvbs = (\rvbb, \rvbzi)$;\\
                
                $\text{ATE}^{\text{r}}_z = \text{ATE}^{\text{r}}_z + \frac{\sum_{j: t_j = 1} \sum_{t'} \Expectation\normalbrackets{\rvy | \rvbz_j,t'} \Probability(t')}{\normalabs{\normalbraces{j: t_j = 1}}} - \frac{\sum_{j: t_j = 0} \sum_{t'} \Expectation\normalbrackets{\rvy | \rvbz_j,t'} \Probability(t')}{\normalabs{\normalbraces{j: t_j = 0}}}$;\\
                $\text{ATE}^{\text{r}}_s = \text{ATE}^{\text{r}}_s + \frac{\sum_{j: t_j = 1} \sum_{t'} \Expectation\normalbrackets{\rvy | \rvbs_j,t'} \Probability(t')}{\normalabs{\normalbraces{j: t_j = 1}}} - \frac{\sum_{j: t_j = 0} \sum_{t'} \Expectation\normalbrackets{\rvy | \rvbs_j,t'} \Probability(t')}{\normalabs{\normalbraces{j: t_j = 0}}}$;\\
                }
            }
        }
    }
        \If{$c_2 > 0$}
        {
        $\text{ATE}_z = \text{ATE}_z + \text{ATE}^{r}_z / c_2$, $\text{ATE}_s = \text{ATE}_s + \text{ATE}^{r}_s / c_2$, $c_1 = c_1 + 1$;
        }
}
\If{$c_1 > 0$}
{
$\text{ATE}_z = \text{ATE}_z / c_1, \text{ATE}_s = \text{ATE}_s / c_1$;
}
\Else
{
Failed to find $\rvbz = (\rvbzi, \rvbzo)$ satisfying \cref{eq_ci,eq_condition};
}
\caption{ATE estimation using subset search.}
\label{alg:subset_search}
\end{algorithm}
The ATE can be computed by taking the first moment version of \cref{eq_thm_main_1} or \cref{eq_thm_main_2}. In Algorithm \ref{alg:subset_search}, we provide a systematic way to estimate the ATE using Theorem \ref{thm_main} by searching for a set $\rvbz \in \cZ \defn \cV \setminus \{\rvt, \rvbb, \rvy\}$ such that (a) p-value of conditional independence in \cref{eq_ci} passes a threshold $p_v$ and (b) there exists a decomposition $\rvbz = (\rvbzi,\rvbzo)$ such that p-values of conditional independencies in \cref{eq_condition} pass the threshold $p_v$. Then, for every such $\rvbz$, the algorithm computes the ATE using the first moment version of \cref{eq_thm_main_1}, and averages. The algorithm produces another estimate by using \cref{eq_thm_main_2} instead of \cref{eq_thm_main_1}.

\subsection{Relation to PAG-based algorithms} 
Now, we exhibit how our approach can recover the causal effect in certain scenarios where PAG-based methods are not suitable. PAGs depict ancestral relationships (not necessarily direct) with directed edges and ambiguity in orientations (if they exist across members of the equivalence class) by circle marks. Figure \ref{fig_pag_example}(c) shows the PAG consistent with SMCM in Figure \ref{fig_pag_example}(a). While we formally define PAGs in Appendix \ref{app_pag}, we refer interested readers to \cite{triantafillou2015constraint}. The IDP algorithm  of \citet{jaber2019causal} is sound and complete for identifying causal effect from PAGs.

Consider SMCM in Figure \ref{fig_pag_example}(a) where our approach recovers the causal effect as $(i)$ Assumptions \ref{assumption_descendant} to \eqref{assumption_known_children}, $(ii)$ \eqref{eq_ci}, and $(iii)$ \eqref{eq_condition} hold (where $(ii)$ and $(iii)$ can be tested from observational data). However, the IDP algorithm fails to recover the effect from the PAG. To see this, consider SMCM in Figure \ref{fig_pag_example}(b) which is Markov equivalent to SMCM in Figure \ref{fig_pag_example}(a), i.e., the PAG in Figure \ref{fig_pag_example}(c) is also consistent with SMCM in Figure \ref{fig_pag_example}(b). Intuitively, when the strength of the edge between $\rvt$ and $\rvbb$ is very small but the strength of the edge between $\rvt$ and $\rvy$ is very high for both Figure \ref{fig_pag_example}(a) and Figure \ref{fig_pag_example}(b), causal effect in Figure \ref{fig_pag_example}(b) remains high while the causal effect in Figure \ref{fig_pag_example}(a)  goes to zero. We note that Assumptions \ref{assumption_descendant} and \ref{assumption_known_children}, and \eqref{eq_ci} do not hold for the SMCM in Figure \ref{fig_pag_example}(b).
\begin{figure}[h]
\vspace{-3mm}
    \centering
    \begin{tabular}{c}
    \begin{tikzpicture}[scale=0.6, every node/.style={transform shape}, > = latex, shorten > = 1pt, shorten < = 1pt]
	\node[shape=circle,draw=black](y) at (6,-1.5) {\LARGE$\rvy$};
	\node[shape=circle,draw=black](ztilde) at (4,-1.5) {\Large$\rvbzo$};
	\node[shape=circle,draw=black](zhat) at (3,0.5) {\Large$\rvbzi$};
	\node[shape=circle,draw=black](b) at (2,-1.5) {\Large$\rvbb$};
	\node[shape=circle,draw=black](t) at (0,-1.5) {\LARGE$\rvt$};
	\path[style=thick][<->, bend right](t) [dashed] edge (y);
	\path[style=thick][<->, bend left](zhat) [dashed] edge (y);
	\path[style=thick][->](zhat) edge (ztilde);
	\path[style=thick][->](t) edge (b);
	\path[style=thick][->](b) edge (ztilde);
	\path[style=thick][->](ztilde) edge (y);
	\path[style=thick][<-](b) edge (zhat);
	\end{tikzpicture} 
     \hspace{3mm}  
    \begin{tikzpicture}[scale=0.6, every node/.style={transform shape}, > = latex, shorten > = 1pt, shorten < = 1pt]
	\node[shape=circle,draw=black](y) at (6,-1.5) {\LARGE$\rvy$};
	\node[shape=circle,draw=black](ztilde) at (4,-1.5) {\Large$\rvbzo$};
	\node[shape=circle,draw=black](zhat) at (3,0.5) {\Large$\rvbzi$};
	\node[shape=circle,draw=black](b) at (2,-1.5) {\Large$\rvbb$};
	\node[shape=circle,draw=black](t) at (-0.5,-1.5) {\LARGE$\rvt$};
	\path[style=thick][->, bend right](t) edge (y);
	\path[style=thick][<->, bend left](zhat) [dashed] edge (y);
	\path[style=thick][->](zhat) edge (ztilde);
	\path[style=thick][<->](t) [dashed] edge (b);
	\path[style=thick][->](b) edge (ztilde);
	\path[style=thick][->](ztilde) edge (y);
	\path[style=thick][<-](b) edge (zhat);
	\end{tikzpicture}
    \hspace{3mm}
     \begin{tikzpicture}[scale=0.6, every node/.style={transform shape}, > = latex, shorten > = 1pt, shorten < = 1pt]
	\node[shape=circle,draw=black](y) at (6,-1.5) {\LARGE$\rvy$};
	\node[shape=circle,draw=black](ztilde) at (4,-1.5) {\Large$\rvbzo$};
	\node[shape=circle,draw=black](zhat) at (3,0.5) {\Large$\rvbzi$};
	\node[shape=circle,draw=black](b) at (2,-1.5) {\Large$\rvbb$};
	\node[shape=circle,draw=black](t) at (0,-1.5) {\LARGE$\rvt$};
	\path[style=thick][{Circle[open]}->, bend right](t) edge (y);
	\path[style=thick][{Circle[open]}->](t) edge (b);
	\path[style=thick][->](b) edge (ztilde);
	\path[style=thick][->](ztilde) edge (y);
	\path[style=thick][<-{Circle[open]}](b) edge (zhat);
	\path[style=thick][<-](ztilde) edge (zhat);
	\path[style=thick][<-{Circle[open]}](y) edge (zhat);
	\end{tikzpicture}\\
    \end{tabular}
    \caption{(a) An SMCM satisfying \eqref{eq_ci} and \eqref{eq_condition}. (b) An SMCM obtained from (a) by modifying the edges between $\rvt$ and $\rvbb$ and between $\rvt$ and $\rvy$. {(c)} The PAG corresponding to SMCM in (a) and (b).} 
    \label{fig_pag_example}
\end{figure}
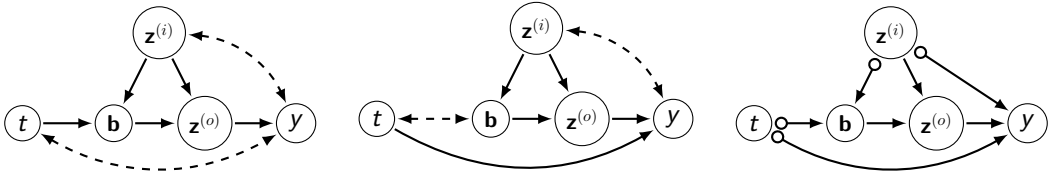
\begin{remark}
    Obtaining a PAG typically requires a large number of conditional independence tests \citep{claassen2013learning} and erroneous tests can potentially alter the structure of the PAG non-locally. Moreover, incorporating arbitrary side information into a PAG in a systematic way is still an open problem. In contrast, our approach does not rely on constructing a graphical object such as a PAG. 
\end{remark}
\section{Empirical Evaluation}
\label{sec_experiments}
We evaluate our approach empirically in 3 ways: $(i)$ we demonstrate the applicability of our method on a class of random graphs, $(ii)$ we assess the effectiveness of our method in estimating the ATE using finite samples, and $(iii)$ we showcase the  potential of our method for causal fairness analysis.
\subsection{Applicability to a class of random graphs}
\label{subsec_random_graphs}
In this experiment, we create a class of random SMCMs, sample 100 SMCMs from this class, and check if \cref{eq_ci,eq_condition} hold by checking for corresponding d-separations in the SMCMs. \\

\noindent \textbf{Creation of random SMCMs.} 
Let $p \defn |\cV|$ denote the dimension of observed variables including $\rvbx$, $\rvt$, and $\rvy$. Let $\rvv_1, \cdots, \rvv_p$ denote a causal ordering of these variables. Our random ensemble depends on two parameters: $(i)$ $d \leq p/2$ which is the expected in-degree of variables $\rvv_{2d}, \cdots, \rvv_p$ and $(ii)$ $q \leq p$ which controls the number of unobserved features. For $1 \leq i < j \leq p$, we add $\rvv_i \myrightarrow \rvv_j$ with probability 0.5 if $j \leq 2d$ and with probability $d/(j-1)$ if $j > 2d$. We note that this procedure is such that the expected in-degree for $\rvv_{2d}, \cdots, \rvv_p$  is $d$ as desired. Next, for $1 \leq i < j \leq p$, we add $\rvv_i \doublearrow \rvv_j$ with probability $q/p$. Then, we choose $\rvv_p$ as $\rvy$, any variable that is ancestor of $\rvy$ but not its parent or grandparent as $\rvt$, and all children of $\rvt$ as $\rvbb$. Finally, we add $\rvt \doublearrow \rvy$ if missing.\\

\noindent \textbf{Results.} We compare the success rate of two approaches: $(i)$ exhaustive search for $\rvbz$ satisfying \cref{eq_ci,eq_condition} which is exponential in $p$ and $(ii)$ search for a $\rvbz$ of size at-most 5 satisfying \cref{eq_ci,eq_condition} which is polynomial in $p$. We provide the number of successes of these approaches as a tuple in Table \ref{table_success_rate_1} for various $p$, $d$, and $q$. We see that the two approaches have comparable performances. We also compare with the IDP algorithm by providing it the true PAG. However, it gives 0 successes across various $p$, $d$, and $q$. We provide results for another random ensemble in Appendix \ref{appendix_experiments}.

\begin{table}[h]
  \caption{Number of successes out of 100 random graphs for our methods shown as a tuple. The first method searches a $\rvbz$ exhaustively and the second method searches a $\rvbz$ with size at-most 5.} 
  \label{table_success_rate_1}
  \vspace{1mm}
\centering
  \begin{tabular}{ccccccccc}
    \toprule
    \multirow{3}{*}{} &
      \multicolumn{3}{c}{$p = 10$} &
      \multicolumn{3}{c}{$p = 15$} \\
      &  $d = 2$ & $d = 3$ & $d = 4$ & $d = 2$ & $d = 3$ & $d = 4$  \\
      \midrule
    \midrule
    $q = 0.0$ & $(43, 43)$ & $(20, 20)$ & $(21, 21)$ & $(27, 26)$ & $(9, 9)$ & $(4, 2)$\\
    $q = 0.5$ & $(23, 23)$ & $(16, 16)$ & $(7, 7)$ & $(18, 17)$ & $(4, 3)$ & $(0, 0)$\\
    $q = 1.0$ & $(6, 6)$ & $(4, 4)$ & $(5, 5)$ & $(9, 9)$ & $(10, 9)$ & $(0, 0)$\\
    \bottomrule
  \end{tabular}
\end{table}
\subsection{Estimating the ATE}
\label{sec:descsynth}
In this experiment, we generate synthetic data using the 6 random SMCMs in Section \ref{subsec_random_graphs} for $p = 10$, $d = 2$, and $q = 1.0$ where our approach was successful indicating existence of $\rvbz = (\rvbzi,\rvbzo)$ such that the conditional independence statements in Theorem \ref{thm_main} hold. Then, we use Algorithm \ref{alg:subset_search} to compute the error in estimating ATE and compare against a \texttt{Baseline} which uses the front-door adjustment in \cref{eqn:front-door} with $\rvbz=\rvbb$ given the side information in Assumption \ref{assumption_known_children}. We provide the results for the same experiment for specific choices of SMCMs including the one in Figure \ref{fig_toy_example} in Appendix \ref{appendix_experiments}. We also provide the 6 random SMCMs in Appendix \ref{appendix_experiments}. We use RCoT hypothesis test \citep{strobl2019approximate} for conditional independence testing from finite data.\\

\noindent \textbf{Data generation.} We use the following procedure to generate data from every SMCM. We generate unobserved variables independently from $\mathrm{Unif}[1,2]$ which denotes the uniform distribution over $[1,2]$. For every observed variable $\rvv \in \cV$, let $\pi(\rvv) \defn (\pao(\rvv), \pau(\rvv)) \in \mathbb{R}^{d_v \times 1}$ denote the set of observed and unobserved parents of $\rvv$ stacked as a column vector. Then, we generate $\rvv \in \cV$ as
\begin{align} \label{syn:func_model}
    \rvv = \mathbf{a}_{v}^{\top} \pi(\rvv) + 0.1 ~~ {\cal N}(0,1) \stext{for} \rvv \in \cV \setminus \{\rvt\} \qtext{and} \rvt = \mathrm{Bernoulli}(\mathrm{Sigmoid}(\mathbf{a}_{t}^{\top} \pi(\rvt)))
\end{align}
where the coefficients $\mathbf{a}_{v} \in \mathbb{R}^{d_v \times 1}$ with every entry sampled independently from $\mathrm{Unif}[1,2]$. Also, to generate the true ATE, we intervene on the generation model in \cref{syn:func_model} by setting $t=0$ and $t = 1$. \\

\begin{figure}[h!]
    \centering
    \includegraphics[width=0.5\linewidth]{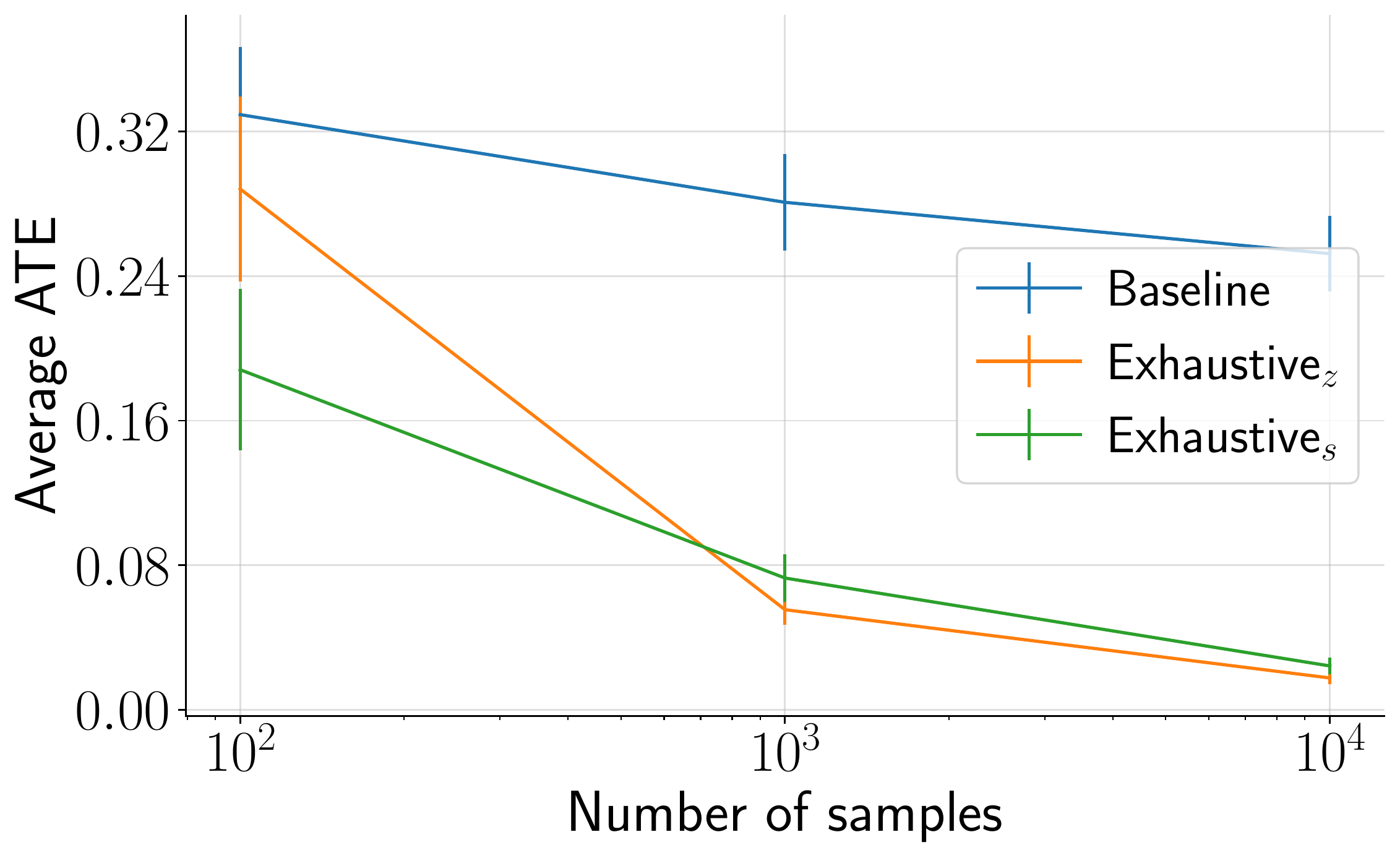}
\caption{Average ATE for Algorithm \ref{alg:subset_search} and \texttt{Baseline} vs. %
number of samples.}
\label{fig_ate_vs_samples}
\end{figure}

\noindent \textbf{Results.} For every SMCM, we generate 
$n$ samples of every observed variable in every run of the experiment. We average the ATE error over 10 such runs where the coefficients in \cref{syn:func_model} vary across runs. We report the average of these averages over the 6 SMCMs in Figure \ref{fig_ate_vs_samples} for various $n$. While the error rates of \texttt{Baseline} and Algorithm \ref{alg:subset_search} are of the similar order for $n = 100$, Algorithm \ref{alg:subset_search} gives much lower errors for $n = 1000$ and $n = 10000$ showing the efficacy of our method.

\subsection{Experiments with real-world fairness benchmarks}
\label{subsec_expts_real_world}
Next, we describe how our results enable finding front-door-like adjustment sets in fairness problems. In a typical fairness problem, the goal is to ensure that the outcome variable $\rvy$ does not unfairly depend on the sensitive/protected attribute, e.g., race or gender (which we define to be treatment variable $\rvt)$, which would reflect undesirable biases. Often, the outcome is a descendant of the sensitive attribute (as per Assumption \ref{assumption_descendant}), and both outcome and sensitive attribute are confounded by unobserved variables (as per Assumption \ref{assumption_confounded}). Furthermore, there are be a multitude of measured post-sensitive-attribute variables that can affect the outcome. This stands in contrast to the usual settings for causal effect estimation, where pre-treatment variables are primarily utilized.

Fairness problems are typically evaluated using various fairness metrics, such as causal fairness metrics or observational metrics. Causal metrics require knowing the underlying causal graph, which can be a challenge in practice. Observational criteria can be decomposed into three types of effects \citep{zhang2018fairness, plecko2022causal}: spurious effects, direct effects, and indirect effects (through descendants of sensitive attribute). In some scenarios, capturing the sum of direct and indirect effects is of interest, but even this requires knowing the causal graph.

Now, we demonstrate the application of our adjustment formulae in Theorem \ref{thm_main} to compute the sum of direct and indirect effects of the sensitive attribute on the outcome, while separating it from spurious effects. The sum of these effects is indeed the causal effect of sensitive attribute on the outcome. In other words, we consider the following fairness metric: $\mathbb{E}[\rvy | do (t=1)] - \mathbb{E}[\rvy | do(t=0)]$. We assume that all the children of the sensitive attribute are known, which may be easier to justify compared to the typical assumption in causal fairness literature of knowing the entire causal graph.

\paragraph{German Credit Dataset.}
The {German Credit dataset} \citep{german_credit_data} is used for credit risk analysis where the goal is to predict whether a loan applicant is a good or bad credit risk based on applicant's 20 demographic and socio-economic attributes. The binary credit risk is the outcome $\rvy$ and the applicant's age (binarized by thresholding at $25$ \citep{kamiran2009classifying}) is the sensitive attribute $\rvt$. Further, the categorical attributes are one-hot encoded.
 
We apply Algorithm \ref{alg:subset_search} with $n_r = 100$ and $p_v = 0.1$ where we search for a set $\rvbz = (\rvbzo, \rvbzi)$ of size at most $3$ under the following two distinct assumptions on the set of all children $\rvbb$ of $\rvt$:
\begin{enumerate}[topsep=0pt,itemsep=0pt]
    \item When considering $\rvbb\!=\! \{$\# of people financially dependent on the applicant, applicant's savings, applicant's job$\}$, Algorithm \ref{alg:subset_search} results in $\rvbzi\!=\! \{$purpose for which the credit was needed, indicator of whether the applicant was a foreign worker$\}$, $\rvbzo\!=\! \{$installment plans from providers other than the credit-giving bank$\}$, $\text{ATE}_z = 0.0125 \pm 0.0011$, and $\text{ATE}_s = 0.0105 \pm 0.0018$.
\item When considering $\rvbb\!=\! \{$\# of people financially dependent on the applicant, applicant's savings$\}$, Algorithm \ref{alg:subset_search} results in $\rvbzi\!=\! \{$purpose for which the credit was needed, applicant’s checking account status with the bank$\}$, $\rvbzo\!=\! \{$installment plans from providers other than the credit-giving bank$\}$, $\text{ATE}_z = 0.0084 \pm 0.0008$, and $\text{ATE}_s = -0.0046 \pm 0.0021$.
\end{enumerate}
\vspace{1mm}
Under the first assumption above, the causal effect using the adjustment formulae in \eqref{eq_thm_main_1} and \eqref{eq_thm_main_2} have same sign and are close in magnitude. However, under the second assumption, the effect flips sign. The results suggest that the second hypothesis regarding $\rvbb$ is incorrect, implying that applicant's job may indeed be a direct child of applicant's age, which aligns with intuition.

\begin{figure}[ht]
\vspace{-3mm}
    \centering
    \includegraphics[width=0.75\linewidth,clip]{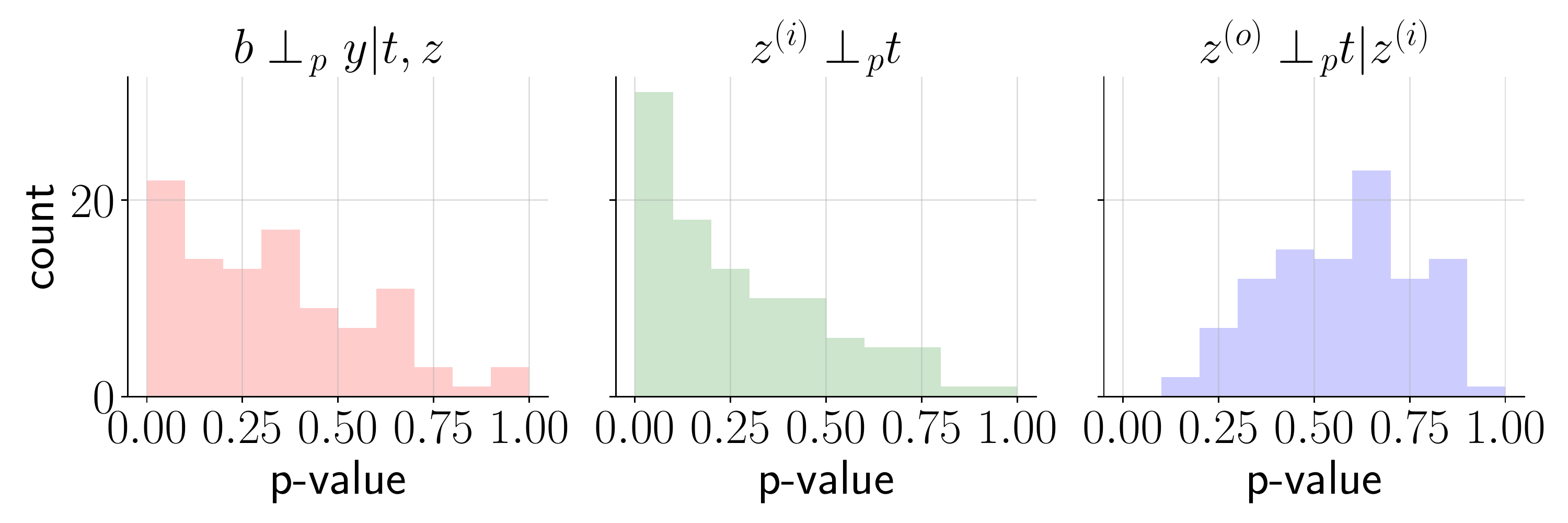}
\caption{Histograms of p-values of the conditional independencies in \eqref{eq_thm_main_1} and \eqref{eq_thm_main_2} over 100 bootstrap runs for $\rvbb\!=\! \{$\# of people financially dependent on the applicant, applicant's savings, applicant's job$\}$.}
\label{fig_german_1}
\end{figure}

The dataset has only 1000 samples, which increases the possibility of detecting independencies in our criterion by chance, even with the size of $\rvbz$ constrained. To address this issue, we use 100 random bootstraps with a sample size equal to half of the training data and evaluate the p-value of our conditional independence criteria for all subsets returned by our algorithm. We select the subset $\rvbz$ with the highest median p-value (computed over the bootstraps) and use it in our adjustment formulae on a held out test set. To assess the conditional independencies associated with the selected $\rvbz$, we plot a histogram of the corresponding p-values for all these bootstraps. If the conditional independencies hold, we expect the p-values to be spread out, which we observe in the histograms in Figure \ref{fig_german_1} for the first choice of $\rvbb$. We report similar results for the second choice of $\rvbb$ in Appendix \ref{appendix_experiments}.

\noindent \textbf{Adult Dataset:}
We perform a similar analysis on the Adult dataset \citep{adult1996}. With suitable choices of $\rvbb$, Algorithm \ref{alg:subset_search} was unable to find a suitable $\rvbz$ satisfying $\rvbb \indep \rvy | \rvbz, \rvt$. This suggests that in this dataset, there may not be any non-child descendants of the sensitive attribute, which is required for our criterion to hold. More details can be found in Appendix \ref{appendix_experiments}.
\section{Conclusion and Discussion}
In this work, we proposed sufficient conditions for causal effect estimation without requiring the knowledge of the entire causal graph using front-door-like adjustment given structural side information. We showed our approach can identify causal effect in graphs where known Markov equivalence classes do not allow identification. 

Our approach relies primarily on two assumptions: Assumption \ref{assumption_confounded} and Assumption \ref{assumption_known_children}. Assumption \ref{assumption_confounded} plays a crucial role in Theorem \ref{thm_main} (see the discussion in Section \ref{app_ass_confounded}) as it requires the presence of an unobserved confounder between the treatment variable and the outcome variable. This assumption is necessary for the applicability of our approach. If Assumption \ref{assumption_confounded} does not hold, it implies that there is a set that satisfies the back-door criterion, and existing methods for finding back-door adjustment sets \citep{entner2013data,cheng2020towards,shah2022finding} can be utilized. However, it is important to note that in many real-world scenarios, the presence of unobserved variables that potentially confound the treatment and the outcome is common, and Assumption \ref{assumption_confounded} holds in such cases. Assumption \ref{assumption_known_children} is the requirement of knowing the entire set of children of the treatment variable within the causal graph. While this is strictly less demanding than specifying the entire causal graph, it may still present practical challenges in real-world scenarios. For instance, in large-scale observational studies or domains with numerous variables, exhaustively identifying all the children may be computationally demanding. Therefore, it is important to understand whether one can estimate the causal effect using front-door-like adjustment with even less side information, e.g., knowing only one child of the treatment variable or knowing any subset of children of the treatment variable. However, until then, one could seek input from domain experts. These experts possess valuable knowledge and insights about the specific domain under study, which can aid in identifying all the relevant variables that serve as children of the treatment.

Lastly, Algorithm \ref{alg:subset_search} has an exponential time complexity due to its search over all possible subsets of observed variables (except $\rvt, \rvbb, \rvy$). While this is inherent in the general case, recent work by \citet{shah2022finding} proposed a scalable approximation for conditional independence testing using continuous optimization  by exploiting the principle of invariant risk minimization, specifically for back-door adjustment without the need for the causal graph. However, extending this approach to multiple conditional independence tests, as required in \cref{eq_ci,eq_condition}, remains an open challenge. Therefore, exploring the development of continuous optimization-based methods for scalability of front-door adjustment in the absence of the causal graph is an exciting direction for future work.

\section*{Acknolwedgements}

Murat Kocaoglu acknowledges the support of NSF Grant CAREER 2239375.
\bibliographystyle{abbrvnat}
\bibliography{main}

\begin{thebibliography}{40}
\providecommand{\natexlab}[1]{#1}
\providecommand{\url}[1]{\texttt{#1}}
\expandafter\ifx\csname urlstyle\endcsname\relax
  \providecommand{\doi}[1]{doi: #1}\else
  \providecommand{\doi}{doi: \begingroup \urlstyle{rm}\Url}\fi

\bibitem[Acharya et~al.(2018)Acharya, Bhattacharyya, Daskalakis, and
  Kandasamy]{acharya2018learning}
J.~Acharya, A.~Bhattacharyya, C.~Daskalakis, and S.~Kandasamy.
\newblock Learning and testing causal models with interventions.
\newblock \emph{Advances in Neural Information Processing Systems}, 31, 2018.

\bibitem[Bellemare et~al.(2019)Bellemare, Bloem, and
  Wexler]{bellemare2019paper}
M.~F. Bellemare, J.~R. Bloem, and N.~Wexler.
\newblock The paper of how: Estimating treatment effects using the front-door
  criterion.
\newblock Technical report, Working paper, 2019.

\bibitem[Bhattacharya and Nabi(2022)]{bhattacharya2022testability}
R.~Bhattacharya and R.~Nabi.
\newblock On testability of the front-door model via verma constraints.
\newblock In \emph{Uncertainty in Artificial Intelligence}, pages 202--212.
  PMLR, 2022.

\bibitem[Castro et~al.(2020)Castro, Walker, and Glocker]{castro2020causality}
D.~C. Castro, I.~Walker, and B.~Glocker.
\newblock Causality matters in medical imaging.
\newblock \emph{Nature Communications}, 11\penalty0 (1):\penalty0 1--10, 2020.

\bibitem[Cheng et~al.(2020)Cheng, Li, Liu, Yu, Lee, and Liu]{cheng2020towards}
D.~Cheng, J.~Li, L.~Liu, K.~Yu, T.~D. Lee, and J.~Liu.
\newblock Towards unique and unbiased causal effect estimation from data with
  hidden variables.
\newblock \emph{arXiv preprint arXiv:2002.10091}, 2020.

\bibitem[Cheng et~al.(2022)Cheng, Li, Liu, Liu, and Le]{cheng2022data}
D.~Cheng, J.~Li, L.~Liu, J.~Liu, and T.~D. Le.
\newblock Data-driven causal effect estimation based on graphical causal
  modelling: A survey.
\newblock \emph{arXiv preprint arXiv:2208.09590}, 2022.

\bibitem[Claassen et~al.(2013)Claassen, Mooij, and
  Heskes]{claassen2013learning}
T.~Claassen, J.~M. Mooij, and T.~Heskes.
\newblock Learning sparse causal models is not np-hard.
\newblock In \emph{Proceedings of the Twenty-Ninth Conference on Uncertainty in
  Artificial Intelligence}, pages 172--181, 2013.

\bibitem[Entner et~al.(2013)Entner, Hoyer, and Spirtes]{entner2013data}
D.~Entner, P.~Hoyer, and P.~Spirtes.
\newblock Data-driven covariate selection for nonparametric estimation of
  causal effects.
\newblock In \emph{Artificial Intelligence and Statistics}, pages 256--264.
  PMLR, 2013.

\bibitem[Fulcher et~al.(2020)Fulcher, Shpitser, Marealle, and
  Tchetgen~Tchetgen]{fulcher2020robust}
I.~R. Fulcher, I.~Shpitser, S.~Marealle, and E.~J. Tchetgen~Tchetgen.
\newblock Robust inference on population indirect causal effects: the
  generalized front door criterion.
\newblock \emph{Journal of the Royal Statistical Society: Series B (Statistical
  Methodology)}, 82\penalty0 (1):\penalty0 199--214, 2020.

\bibitem[Glynn and Kashin(2017)]{glynn2017front}
A.~N. Glynn and K.~Kashin.
\newblock Front-door difference-in-differences estimators.
\newblock \emph{American Journal of Political Science}, 61\penalty0
  (4):\penalty0 989--1002, 2017.

\bibitem[Glynn and Kashin(2018)]{glynn2018front}
A.~N. Glynn and K.~Kashin.
\newblock Front-door versus back-door adjustment with unmeasured confounding:
  Bias formulas for front-door and hybrid adjustments with application to a job
  training program.
\newblock \emph{Journal of the American Statistical Association}, 113\penalty0
  (523):\penalty0 1040--1049, 2018.

\bibitem[Gultchin et~al.(2020)Gultchin, Kusner, Kanade, and
  Silva]{gultchin2020differentiable}
L.~Gultchin, M.~Kusner, V.~Kanade, and R.~Silva.
\newblock Differentiable causal backdoor discovery.
\newblock In \emph{International Conference on Artificial Intelligence and
  Statistics}, pages 3970--3979. PMLR, 2020.

\bibitem[Gupta et~al.(2021)Gupta, Lipton, and Childers]{gupta2021estimating}
S.~Gupta, Z.~C. Lipton, and D.~Childers.
\newblock Estimating treatment effects with observed confounders and mediators.
\newblock In \emph{Uncertainty in Artificial Intelligence}, pages 982--991.
  PMLR, 2021.

\bibitem[Hofmann(1994)]{german_credit_data}
H.~Hofmann.
\newblock {Statlog (German Credit Data)}.
\newblock UCI Machine Learning Repository, 1994.

\bibitem[H{\"u}nermund and Bareinboim(2019)]{hunermund2019causal}
P.~H{\"u}nermund and E.~Bareinboim.
\newblock Causal inference and data fusion in econometrics.
\newblock \emph{arXiv preprint arXiv:1912.09104}, 2019.

\bibitem[Jaber et~al.(2019)Jaber, Zhang, and Bareinboim]{jaber2019causal}
A.~Jaber, J.~Zhang, and E.~Bareinboim.
\newblock Causal identification under markov equivalence: Completeness results.
\newblock In \emph{International Conference on Machine Learning}, pages
  2981--2989. PMLR, 2019.

\bibitem[Jeong et~al.(2022)Jeong, Tian, and Bareinboim]{jeong2022finding}
H.~Jeong, J.~Tian, and E.~Bareinboim.
\newblock Finding and listing front-door adjustment sets.
\newblock \emph{arXiv preprint arXiv:2210.05816}, 2022.

\bibitem[Kamiran and Calders(2009)]{kamiran2009classifying}
F.~Kamiran and T.~Calders.
\newblock Classifying without discriminating.
\newblock In \emph{2009 2nd international conference on computer, control and
  communication}, pages 1--6. IEEE, 2009.

\bibitem[Kohavi and Becker(1996)]{adult1996}
R.~Kohavi and B.~Becker.
\newblock {UCI} machine learning repository, 1996.
\newblock URL \url{http://archive.ics.uci.edu/ml/datasets/adult}.

\bibitem[Kuroki(2000)]{kuroki2000selection}
M.~Kuroki.
\newblock Selection of post-treatment variables for estimating total effect
  from empirical research.
\newblock \emph{Journal of the Japan Statistical Society}, 30\penalty0
  (2):\penalty0 115--128, 2000.

\bibitem[Kuroki and Cai(2012)]{kuroki2012selection}
M.~Kuroki and Z.~Cai.
\newblock Selection of identifiability criteria for total effects by using path
  diagrams.
\newblock \emph{arXiv preprint arXiv:1207.4140}, 2012.

\bibitem[Mogstad and Torgovitsky(2018)]{mogstad2018identification}
M.~Mogstad and A.~Torgovitsky.
\newblock Identification and extrapolation of causal effects with instrumental
  variables.
\newblock \emph{Annual Review of Economics}, 10:\penalty0 577--613, 2018.

\bibitem[Nabi et~al.(2019)Nabi, Malinsky, and Shpitser]{nabi2019learning}
R.~Nabi, D.~Malinsky, and I.~Shpitser.
\newblock Learning optimal fair policies.
\newblock In \emph{International Conference on Machine Learning}, pages
  4674--4682. PMLR, 2019.

\bibitem[Pearl(1993)]{Pearl1993}
J.~Pearl.
\newblock [bayesian analysis in expert systems]: Comment: graphical models,
  causality and intervention.
\newblock \emph{Statistical Science}, 8\penalty0 (3):\penalty0 266--269, 1993.

\bibitem[Pearl(1995)]{pearl1995causal}
J.~Pearl.
\newblock Causal diagrams for empirical research.
\newblock \emph{Biometrika}, 82\penalty0 (4):\penalty0 669--688, 1995.

\bibitem[Pearl(2009)]{pearl2009causality}
J.~Pearl.
\newblock \emph{Causality}.
\newblock Cambridge university press, 2009.

\bibitem[Perkovic et~al.(2018)Perkovic, Textor, Kalisch, and
  Maathuis]{perkovic2018complete}
E.~Perkovic, J.~Textor, M.~Kalisch, and M.~H. Maathuis.
\newblock Complete graphical characterization and construction of adjustment
  sets in markov equivalence classes of ancestral graphs.
\newblock \emph{The Journal of Machine Learning Research}, 2018.

\bibitem[Plecko and Bareinboim(2022)]{plecko2022causal}
D.~Plecko and E.~Bareinboim.
\newblock Causal fairness analysis.
\newblock \emph{arXiv preprint arXiv:2207.11385}, 2022.

\bibitem[Shah et~al.(2022)Shah, Shanmugam, and Ahuja]{shah2022finding}
A.~Shah, K.~Shanmugam, and K.~Ahuja.
\newblock Finding valid adjustments under non-ignorability with minimal dag
  knowledge.
\newblock In \emph{International Conference on Artificial Intelligence and
  Statistics}, pages 5538--5562. PMLR, 2022.

\bibitem[Shpitser and Pearl(2006)]{shpitser2006identification}
I.~Shpitser and J.~Pearl.
\newblock Identification of joint interventional distributions in recursive
  semi-markovian causal models.
\newblock In \emph{Proceedings of the National Conference on Artificial
  Intelligence}, volume~21, page 1219. Menlo Park, CA; Cambridge, MA; London;
  AAAI Press; MIT Press; 1999, 2006.

\bibitem[Spirtes et~al.(2000)Spirtes, Glymour, and
  Scheines]{spirtes2000causation}
P.~Spirtes, C.~N. Glymour, and R.~Scheines.
\newblock \emph{Causation, prediction, and search}.
\newblock MIT press, 2000.

\bibitem[Strobl et~al.(2019{\natexlab{a}})Strobl, Spirtes, and
  Visweswaran]{strobl2019estimating}
E.~V. Strobl, P.~L. Spirtes, and S.~Visweswaran.
\newblock Estimating and controlling the false discovery rate of the pc
  algorithm using edge-specific p-values.
\newblock \emph{ACM Transactions on Intelligent Systems and Technology (TIST)},
  10\penalty0 (5):\penalty0 1--37, 2019{\natexlab{a}}.

\bibitem[Strobl et~al.(2019{\natexlab{b}})Strobl, Zhang, and
  Visweswaran]{strobl2019approximate}
E.~V. Strobl, K.~Zhang, and S.~Visweswaran.
\newblock Approximate kernel-based conditional independence tests for fast
  non-parametric causal discovery.
\newblock \emph{Journal of Causal Inference}, 7\penalty0 (1),
  2019{\natexlab{b}}.

\bibitem[Tian and Pearl(2002)]{tian2002general}
J.~Tian and J.~Pearl.
\newblock \emph{A general identification condition for causal effects}.
\newblock eScholarship, University of California, 2002.

\bibitem[Triantafillou and Tsamardinos(2015)]{triantafillou2015constraint}
S.~Triantafillou and I.~Tsamardinos.
\newblock Constraint-based causal discovery from multiple interventions over
  overlapping variable sets.
\newblock \emph{The Journal of Machine Learning Research}, 16\penalty0
  (1):\penalty0 2147--2205, 2015.

\bibitem[Veitch and Zaveri(2020)]{veitch2020sense}
V.~Veitch and A.~Zaveri.
\newblock Sense and sensitivity analysis: Simple post-hoc analysis of bias due
  to unobserved confounding.
\newblock \emph{Advances in Neural Information Processing Systems},
  33:\penalty0 10999--11009, 2020.

\bibitem[Verma and Pearl(1990)]{verma1990causal}
T.~Verma and J.~Pearl.
\newblock Causal networks: Semantics and expressiveness.
\newblock In \emph{Machine intelligence and pattern recognition}, volume~9,
  pages 69--76. Elsevier, 1990.

\bibitem[Wien{\"o}bst et~al.(2022)Wien{\"o}bst, van~der Zander, and
  Li{\'s}kiewicz]{wienobst2022finding}
M.~Wien{\"o}bst, B.~van~der Zander, and M.~Li{\'s}kiewicz.
\newblock Finding front-door adjustment sets in linear time.
\newblock \emph{arXiv preprint arXiv:2211.16468}, 2022.

\bibitem[Zhang(2008)]{zhang2008completeness}
J.~Zhang.
\newblock On the completeness of orientation rules for causal discovery in the
  presence of latent confounders and selection bias.
\newblock \emph{Artificial Intelligence}, 172\penalty0 (16-17):\penalty0
  1873--1896, 2008.

\bibitem[Zhang and Bareinboim(2018)]{zhang2018fairness}
J.~Zhang and E.~Bareinboim.
\newblock Fairness in decision-making—the causal explanation formula.
\newblock In \emph{Proceedings of the AAAI Conference on Artificial
  Intelligence}, volume~32, 2018.

\end{thebibliography}
\clearpage
\appendix
\noindent {\bf \LARGE{Appendix}}
\section{Preliminaries about ancestral graphs}
\label{app_pag}
In this section, we provide the definition of \textit{partial ancestral graphs} (PAGs). PAGs are defined using \textit{maximal ancestral graphs} (MAGs). Below, we define MAGs and PAGs based on their construction from directed acyclic graphs (DAGs).

A MAG can be obtained from a DAG as follows: if two observed nodes $\rvx_1$ and $\rvx_2$ cannot be d-separated conditioned on any subset of observed variables, then $(i)$ $\rvx_1 \myrightarrow \rvx_2$ is added in the MAG if $\rvx_1$ is an ancestor of $\rvx_2$ in the DAG, $(ii)$ $\rvx_2 \myrightarrow \rvx_1$ is added in the MAG if $\rvx_2$ is an ancestor of $\rvx_1$ in the DAG, and $(iii)$ $\rvx_1 \doublearrow \rvx_2$ is added in the MAG if $\rvx_1$ and $\rvx_2$ are not ancestrally related in the DAG. $(iv)$ After the above three operations, if both $\rvx_1 \doublearrow \rvx_2$ and $\rvx_1 \myrightarrow \rvx_2$ are present, we retain only the directed edge. In general, a MAG represents a collection of DAGs that share the same set of observed variables and exhibit the same independence and ancestral relations among these observed variables. It is possible for different MAGs to be Markov equivalent, meaning they represent the exact same independence model.

A PAG shares the same adjacencies as any MAG in the observational equivalence class of MAGs. An end of an edge in the PAG is marked with an arrow ($>$ or $<$) if the edge appears with the same arrow in all MAGs in the equivalence class. An end of an edge in the PAG is marked with a circle ($o$) if the edge appears as an arrow ($>$ or $<$) and a tail ($-$) in two different MAGs in the equivalence class.

\section{Rules of do-calculus}\label{supp:docalculus}
In this section, we provide the do-calculus rules of \citet{pearl1995causal} that are used to prove our main results in the following sections. We build upon the definition of semi-Markovian causal model from Section \ref{sec_prob_formulation}.

For any $\rvbv \in \cW$, let $\cG_{\brvbv}$ be the graph obtained by removing the edges going into $\rvbv$ in $\cG$, and let $\cG_{\urvbv}$ be the graph obtained by removing the edges going out of $\rvbv$ in $\cG$. 
\newcommand{\rules}{Rules of do-calculus}
\begin{theorem}[\rules, \citet{pearl1995causal}]
For any disjoint subsets $\rvbv_1, \rvbv_2, \rvbv_3, \rvbv_4 \subseteq \cW$, we have the following rules.
\begin{enumerate}[label=Rule \arabic*, leftmargin=15mm, labelsep=-0.05em]
    \item: \label{item:rule_1} $\Probability(\rvbv_1 | do(\rvbv_2), \rvbv_3, \rvbv_4) = \Probability(\rvbv_1 | do(\rvbv_2), \rvbv_3)$ $\qquad$ \hspace{1.1cm} if $\rvbv_1 \dsep \rvbv_4 | \rvbv_2, \rvbv_3$ in $\cG_{\brvbv_2}$.
    \item: \label{item:rule_2} $\Probability(\rvbv_1 | do(\rvbv_2), \rvbv_3, do(\rvbv_4)) = \Probability(\rvbv_1 | do(\rvbv_2), \rvbv_3, \rvbv_4)$ $\qquad$ if $\rvbv_1 \dsep \rvbv_4 | \rvbv_2, \rvbv_3$ in $\cG_{\brvbv_2, \urvbv_4}$.
    \item: \label{item:rule_3} $\Probability(\rvbv_1 | do(\rvbv_2), \rvbv_3, do(\rvbv_4)) = \Probability(\rvbv_1 | do(\rvbv_2), \rvbv_3)$ $\qquad$ \hspace{0.4cm} if $\rvbv_1 \dsep \rvbv_4 | \rvbv_2, \rvbv_3$ in $\cG_{\brvbv_2, \wbar{\rvbv_4(\rvbv_3)}}$,
\end{enumerate}
where $\rvbv_4(\rvbv_3)$ is the set of nodes in $\rvbv_4$ that are not ancestors of any node in $\rvbv_3$ in $\cG_{\brvbv_2}$. \citet{pearl1995causal} also gave an alternative criterion for \cref{item:rule_3}.
\begin{enumerate}[label=Rule \arabic*a, leftmargin=17mm, labelsep=-0.05em]
    \setcounter{enumi}{2}
    \item: \label{item:rule_3a} $\Probability(\rvbv_1 | do(\rvbv_2), \rvbv_3, do(\rvbv_4)) = \Probability(\rvbv_1 | do(\rvbv_2), \rvbv_3)$ $\qquad$ \hspace{0.2cm} if $\rvbv_1 \dsep F_{\rvbv_4} | \rvbv_2, \rvbv_3$ in $\cG_{\brvbv_2}^{\rvbv_4}$,
\end{enumerate}
where $\cG^{\rvbv_4}$ is the graph obtained from $\cG$ after adding $(a)$ a node $F_{\rvbv_4}$ and $(b)$ edges from $F_{\rvbv_4}$ to every node in $\rvbv_4$.
\end{theorem}
\noindent Also, throughout our proofs, we use the following fact. 
\begin{fact}\label{fact}
Consider any $\cG'$ obtained by removing any edge(s) from $\cG$. For any sets of variables $\rvbv_1, \rvbv_2, \rvbv_3 \subseteq \cW$, if $\rvbv_1$ and $\rvbv_2$ are d-separated by $\rvbv_3$ in $\cG$ than $\rvbv_1$ and $\rvbv_2$ are d-separated by $\rvbv_3$ in $\cG'$.
\end{fact}

\section{\id}
\label{sec_proof_thm_tian}
In this section, we derive the causal effect for the SMCM in Figure \ref{fig:smcm_fail_condition}(top), i.e., \cref{eq_diff_formula}, as well as prove \cref{thm_tian} one by one.

\subsection{Proof of \cref{eq_diff_formula}}
First, using the law of total probability, we have
\begin{align}
    \Probability(\rvy|\doot) & =  \sum_{z_1, z_2} \Probability(\rvy| \doot, \rvz_1 \!=\! z_1, \rvz_2 \!=\! z_2) \Probability(\rvz_1 \!=\! z_1, \rvz_2 \!=\! z_2| \doot). \label{eq_y_t_marginalized_6}
\end{align}
Now, we show that the  two terms in RHS of \cref{eq_y_t_marginalized_6} can be simplified as follows
\begin{align}
    \Probability(\rvy| \doot, \rvz_1 \!=\! z_1, \rvz_2 \!=\! z_2) & = \sum_{t'} \Probability(\rvy| \rvz_1 \!=\! z_1, \rvz_2 \!=\! z_2, \rvt = t') \Probability(\rvt = t' | \rvz_1 \!=\! z_1). \label{eq_first_term_6}\\
    \Probability(\rvz_1 \!=\! z_1, \rvz_2 \!=\! z_2 | \doot) & = \! \sum_{\svbb} \! \Probability(\rvz_2 \!=\! z_2 | \rvbb = \svbb) \Probability(\rvbb = \svbb | \rvt =t, \rvz_1 \!=\! z_1) \Probability(\rvz_1 \!=\! z_1), \label{eq_second_term_6}
\end{align}
Combining \cref{eq_y_t_marginalized_6,eq_first_term_6,eq_second_term_6} results in \cref{eq_diff_formula}.
\paragraph{Proof of \cref{eq_first_term_6}:}
We have
\begin{align}
    & \Probability(\rvy| \doot, \rvz_1 \!=\! z_1, \rvz_2 \!=\! z_2) \\
    &  \sequal{(a)} \Probability(\rvy = y | \doot, \rvz_1 \!=\! z_1, \rvz_2 \!=\! z_2, \rvbb = \svbb) \\
    & \sequal{(b)} \Probability(\rvy = y | \doot, \rvz_1 \!=\! z_1, \rvz_2 \!=\! z_2, \doob) \\
    & \sequal{(c)} \Probability(\rvy = y | \rvz_1 \!=\! z_1, \rvz_2 \!=\! z_2, \doob)  \\
    & \sequal{(d)} \! \sum_{t'} \! \Probability(\rvy = y | \rvz_1 \!=\! z_1, \rvz_2 \!=\! z_2, \doob, \rvt = t') \Probability(\rvt = t'| \rvz_1 \!=\! z_1, \rvz_2 \!=\! z_2, \doob)\\
    & \sequal{(e)} \! \sum_{t'} \! \Probability(\rvy = y | \rvz_1 \!=\! z_1, \rvz_2 \!=\! z_2, \rvt = t') \Probability(\rvt = t'| \rvz_1 \!=\! z_1, \rvz_2 \!=\! z_2, \doob)\\
    & \sequal{(f)} \! \sum_{t'} \! \Probability(\rvy = y | \rvz_1 \!=\! z_1, \rvz_2 \!=\! z_2, \rvt = t') \Probability(\rvt = t'| \rvz_1 \!=\! z_1, \doob)\\
    & \sequal{(g)} \! \sum_{t'} \! \Probability(\rvy = y | \rvz_1 \!=\! z_1, \rvz_2 \!=\! z_2, \rvt = t') \Probability(\rvt = t'| \rvz_1 \!=\! z_1), \label{eq_y_z_marginalized_6}
\end{align}
where $(a)$ and $(f)$ follow from \cref{item:rule_1},  
$(b)$ follows from \cref{item:rule_2}, $(c)$, $(e)$, and $(g)$ follow from \cref{item:rule_3a}, and $(d)$ follows from the law of total probability.

\paragraph{Proof of \cref{eq_second_term_6}:}
From the law of total probability, we have
\begin{align}
    & \Probability(\rvz_1 \!=\! z_1, \rvz_2 \!=\! z_2 | \doot) \\
    &  = \sum_{\svbb} \Probability(\rvz_1 \!=\! z_1, \rvz_2 \!=\! z_2 | \doot, \rvbb = \svbb) \Probability(\rvbb = \svbb| \doot) \\
    & \sequal{(a)} \sum_{\svbb} \Probability(\rvz_2 \!=\! z_2  | \doot, \rvbb = \svbb)  \Probability(\rvz_1 \!=\! z_1 | \doot, \rvbb = \svbb, \rvz_2 \!=\! z_2 ) \Probability(\rvbb = \svbb| \doot) \\
    & \sequal{(b)} \sum_{\svbb} \Probability(\rvz_2 \!=\! z_2  | \rvbb = \svbb)  \Probability(\rvz_1 \!=\! z_1 | \doot, \rvbb = \svbb, \rvz_2 \!=\! z_2 ) \Probability(\rvbb = \svbb| \doot) \\
    & \sequal{(c)} \sum_{\svbb} \Probability(\rvz_2 \!=\! z_2  | \rvbb = \svbb)  \Probability(\rvz_1 \!=\! z_1 | \doot, \rvbb = \svbb ) \Probability(\rvbb = \svbb| \doot) \\
    & \sequal{(d)}  \sum_{\svbb}\Probability(\rvz_2 \!=\! z_2  | \rvbb = \svbb)  \Probability(\rvz_1 \!=\! z_1, \rvbb = \svbb  | \doot) \\
    & \sequal{(e)} \sum_{\svbb} \Probability(\rvz_2 \!=\! z_2  | \rvbb = \svbb) \Probability(\rvz_1 \!=\! z_1  | \doot) \Probability(\rvbb = \svbb  | \doot, \rvz_1 \!=\! z_1) \\
    & \sequal{(f)} \sum_{\svbb} \Probability(\rvz_2 \!=\! z_2  | \rvbb = \svbb) \Probability(\rvz_1 \!=\! z_1) \Probability(\rvbb = \svbb  | \doot, \rvz_1 \!=\! z_1) \\
    & \sequal{(g)} \sum_{\svbb} \Probability(\rvz_2 \!=\! z_2  | \rvbb = \svbb) \Probability(\rvz_1 \!=\! z_1) \Probability(\rvbb = \svbb  | \rvt = t, \rvz_1 \!=\! z_1)
    \label{eq_z_split_indep_parts_6}
\end{align}
where $(a)$, $(d)$, and $(e)$ follow from the definition of conditional probability, $(b)$ and $(f)$ follows from \cref{item:rule_3a}, $(c)$ follows from \cref{item:rule_1}, and $(g)$ follows from \cref{item:rule_2}.

\subsection{Proof of \cref{thm_tian}}
Let $\mathrm{An}(\rvy)$ denote the union of $\rvy$ and the set of ancestors of $\rvy$, and let $\cG^{\mathrm{An}(\rvy)}$ denote the subgraph of $\cG$ composed only of nodes in $\mathrm{An}(\rvy)$. First, we show that if $\rvbb \dsep \rvy | \rvt, \rvbz$ holds for some $\rvbz$, then there is no bi-directed path between $\rvt$ to $\rvbb$ in $\cG^{\mathrm{An}(\rvy)}$.
\begin{lemma}\label{lemma_tian}
Suppose \cref{assumption_descendant,assumption_confounded,assumption_known_children} hold. Suppose there exists a set $\rvbz  \subseteq \cV \setminus \{\rvt, \rvbb, \rvy\}$ such that $\rvbb \dsep \rvy | \rvt, \rvbz$. Then, there is no bi-directed path between $\rvt$ and $\rvbb$ in $\cG^{\mathrm{An}(\rvy)}$.
\end{lemma}
\noindent Given this claim, \cref{thm_tian} follows from \citet[Theorem 4]{tian2002general}. It remains to prove \cref{lemma_tian}.

\paragraph{Proof of \cref{lemma_tian}.} We prove this result by contradiction. First, from \cref{assumption_descendant,assumption_known_children}, $\rvt \in \mathrm{An}(\rvy)$ and $\rvbb_0 \in \mathrm{An}(\rvy)$ for some $\rvbb_0 \subset \rvbb$. Assume there exists a bi-directed path between $\rvt$ and some $\rvb \in \rvbb_0$ in $\cG^{\mathrm{An}(\rvy)}$. Let $\cP(\rvt, \rvb)$ denote the shortest of these paths. This path is of the form $\rvt \doublearrow \rvv_1 \doublearrow \cdots \doublearrow \rvv_r \doublearrow \rvb$ for some $r \geq 0$ where $\rvv_q \in \cG^{\mathrm{An}(\rvy)}$ for every $q \in [r]$. We have the following two cases depending on the value of $r$.
\begin{enumerate}
    \item[(i)] $r = 0$:  In this case, consider the path $\cP(\rvy, \rvb) \supset \cP(\rvt, \rvb)$ in $\cG$ of the form: $\rvy \doublearrow \rvt \doublearrow \rvb$ in $\cG$ (such a path exists because of \cref{assumption_confounded}). The path $\cP(\rvy, \rvb)$ is unblocked when $\rvt$ and $\rvbz$ are conditioned on contradicting $\rvbb \dsep \rvy | \rvt, \rvbz$.
    \item[(ii)] $r \geq 1$: In this case, consider the path $\cP(\rvy, \rvb) \supset \cP(\rvt, \rvb)$  in $\cG$ of the form: $\rvy \doublearrow \rvt \doublearrow \rvv_1 \doublearrow \cdots \doublearrow \rvv_r \doublearrow \rvb$ (such a path exists because of \cref{assumption_confounded}). We have the following two scenarios depending on whether the path $\cP(\rvy, \rvb)$ is unblocked or blocked when $\rvt$ and $\rvbz$ are conditioned on. Suppose we condition on $\rvt$ and $\rvbz$.
    \begin{enumerate}
        \item The path $\cP(\rvy, \rvb)$ is unblocked: In this case, by assumption, $\rvbb \dsep \rvy | \rvt, \rvbz$ is contradicted.
        \item The path $\cP(\rvy, \rvb)$ is blocked: We create a set $\rvbw$ such that for any $\rvw \in \rvbw$ the following are true: 
        $(a)$ $\rvw = \rvv_q$ for some $q \in [r]$, $(b)$ $\rvw \notin \rvbz$, $(b)$ there is no descendant path $\cP(\rvw, \rvz)$ between $\rvw$ and some $\rvz \in \rvbz$, and $(c)$ there is no descendant path $\cP(\rvw, \rvt)$ between $\rvw$ and $\rvt$.      
        
        In this scenario, $\rvbw \neq \emptyset$ because $\cP(\rvy, \rvb)$ is blocked. Let $\rvw_c \in \rvbw$ be that node which is closest to $\rvb$ in the path $\cP(\rvy, \rvb)$. By the choice of $\rvw_c$, the path $\cP(\rvw_c, \rvb) \subset \cP(\rvy, \rvb)$ is unblocked (when $\rvt$ and $\rvbz$ are conditioned on). Furthermore, by the definition of $\rvbw$, $(a)$ $\rvw_c \in \cG^{\mathrm{An}(\rvy)}$ (because $\rvw_c = \rvv_q$ for some $q \in [r]$) and $(b)$ there exists a descendant path $\cP(\rvw_c, \rvy)$ between $\rvw_c$ and $\rvy$ such that $\rvt \notin \cP(\rvw_c, \rvy)$ as well as $\rvz \notin \cP(\rvw_c, \rvy)$ for every $\rvz \in \rvbz$. Therefore, the path $\cP(\rvw_c, \rvy)$ is unblocked (when $\rvt$ and $\rvbz$ are conditioned on). 
        
        Consider the path $\cP'(\rvy, \rvb)$ obtained after concatenating $\cP(\rvw_c, \rvy)$ and $\cP(\rvw_c, \rvb)$ at $\rvw_c$. This path is unblocked (when $\rvt$ and $\rvbz$ are conditioned on) because: $(a)$ $\cP(\rvw_c, \rvb)$ is unblocked, $(b)$ $\cP(\rvw_c, \rvy)$ is unblocked, and $(c)$ there is no collider at $\rvw_c$ in this path (because $\cP(\rvw_c, \rvy)$ is a descendant path to $\rvy$). However, this contradicts $\rvbb \dsep \rvy | \rvt, \rvbz$.
    \end{enumerate}
\end{enumerate}

\section{\gfd}
\label{sec_proof_thm_main}
In this section, we prove \cref{thm_main}. We begin by stating a few d-separation statements used in this proof. See \cref{sec_proof_lemma_supp_dsep} for a proof.
\begin{lemma}\label{lemma_supp_dsep}
Suppose \cref{assumption_descendant,assumption_confounded,assumption_known_children} and d-separation criteria in Theorem \ref{thm_main}, i.e., \cref{eq_ci,eq_condition}, hold. Then,
\begin{enumerate}[label=(\alph*), itemsep=0.5mm]
    \item \label{item:1} $\rvy \dsep F_{\rvt} | \rvbz, \rvbb$ in $\cG^{\rvt}_{\brvbb}$ and $\rvy \dsep F_{\rvt} | \rvbzi, \rvbb$ in $\cG^{\rvt}_{\brvbb}$,
    \item \label{item:3} $\rvt \dsep \rvbb$ in $\cG_{\urvt}$,  
    \item \label{item:4} $\rvt \dsep \rvbzi | \rvbb$ in $\cG_{\urvt}$, 
    \item \label{item:5} $\rvt \dsep F_{\rvbb} | \rvbzi$ in $\cG^{\rvbb}$, and
    \item \label{item:6} $\rvy \dsep \rvbb | \rvt, \rvbzi$ in $\cG_{\urvbb}$.
\end{enumerate}
\end{lemma}

\noindent Now, we proceed with the proof in two parts. In the first part, we prove \cref{eq_thm_main_1}, and in the second part, we prove \cref{eq_thm_main_2}.

\subsection{Proof of \cref{eq_thm_main_1}}
 First, using the law of total probability, we have
\begin{align}
    \Probability(\rvy = y |\doot) & =  \sum_{\svbz} \Probability(\rvy = y | \doot, \rvbz = \svbz) \Probability(\rvbz = \svbz| \doot). \label{eq_y_t_marginalized}
\end{align}
Now, we show that the  two terms in RHS of \cref{eq_y_t_marginalized} can be simplified as follows
\begin{align}
    \Probability(\rvy = y | \doot, \rvbz = \svbz) & = \sum_{t'} \Probability(\rvy = y | \rvbz = \svbz, \rvt = t') \Probability(\rvt = t'). \label{eq_first_term}\\
    \Probability(\rvbz = \svbz | \doot) & =  \Probability(\rvbz = \svbz | \rvt = t), \label{eq_second_term}
\end{align}
Combining \cref{eq_first_term} and \cref{eq_second_term} completes the proof of \cref{eq_thm_main_1}.
\paragraph{Proof of \cref{eq_first_term}:}
We have
\begin{align}
    \Probability(\rvy = y | \doot, \rvbz = \svbz) &  \sequal{(a)} \Probability(\rvy = y | \doot, \rvbz = \svbz, \rvbb = \svbb) \\
    & \sequal{(b)} \Probability(\rvy = y | \doot, \rvbz = \svbz, \doob) \\
    & \sequal{(c)} \Probability(\rvy = y | \rvbz = \svbz, \doob)  \\
    & \sequal{(d)}  \sum_{t'} \Probability(\rvy = y | \rvbz = \svbz, \doob, \rvt = t') \Probability(\rvt = t'| \rvbz = \svbz, \doob), \label{eq_y_z_marginalized}
\end{align}
where $(a)$ follows from \cref{item:rule_1}, \cref{eq_ci}, and \cref{fact},  $(b)$ follows from \cref{item:rule_2}, \cref{eq_ci}, and \cref{fact}, and $(c)$ follows from \cref{item:rule_3a} and \cref{lemma_supp_dsep}\cref{item:1}, and $(d)$ follows from the law of total probability. \\

\noindent Now, we simplify the first term in \cref{eq_y_z_marginalized} as follows:
\begin{align}
    \Probability(\rvy = y | \rvbz = \svbz, \doob, \rvt = t') & \sequal{(a)} \Probability(\rvy = y | \rvbz = \svbz, \rvbb = \svbb, \rvt = t') \\
    & \sequal{\cref{eq_ci}} \Probability(\rvy = y | \rvbz = \svbz, \rvt = t'), \label{eq_y_z_marginalized_0}
\end{align}
where $(a)$ follows from \cref{item:rule_2}, \cref{eq_ci}, and \cref{fact}. Likewise, we simplify the second term in \cref{eq_y_z_marginalized} as follows:
\begin{align}
    \Probability(\rvt = t'| \doob, \rvbz = \svbz)  \sequal{(a)} \Probability(\rvt = t'| \doob, \rvbzi = \svbzi) & \sequal{(b)} \Probability(\rvt = t'| \rvbzi = \svbzi) \\
    & \sequal{(c)} \Probability(\rvt = t'), \label{eq_y_z_marginalized_1}
\end{align}
where $(a)$ follows from \cref{item:rule_1}, \cref{eq_condition}, and \cref{fact}, $(b)$ follows from \cref{item:rule_3a} and \cref{lemma_supp_dsep}\cref{item:5}, and $(c)$ follows \cref{eq_condition}.\\

\noindent Putting together \cref{eq_y_z_marginalized,eq_y_z_marginalized_0,eq_y_z_marginalized_1} results in \cref{eq_first_term}.

\paragraph{Proof of \cref{eq_second_term}:}
From the law of total probability, we have
\begin{align}
    \Probability(\rvbz = \svbz | \doot) & = \sum_{\svbb} \Probability(\rvbz = \svbz | \doot, \rvbb = \svbb) \Probability(\rvbb = \svbb| \doot). \label{eq_z_t_marginalized}
\end{align}
Now, we simplify the first term in \cref{eq_z_t_marginalized} as follows:
\begin{align}
    & \Probability(\rvbz = \svbz | \doot, \rvbb = \svbb)\\
    & \sequal{(a)} \Probability(\rvbzi = \svbzi | \doot, \rvbb = \svbb) \cdot \Probability(\rvbzo = \svbzo | \doot, \rvbb = \svbb, \rvbzi = \svbzi) \\
    & \sequal{(b)} \Probability(\rvbzi = \svbzi | \doot, \rvbb = \svbb) \cdot \Probability(\rvbzo = \svbzo | \rvt = t, \rvbb = \svbb, \rvbzi = \svbzi)\\
    & \sequal{(c)} \Probability(\rvbzi = \svbzi | \rvt = t, \rvbb = \svbb) \cdot \Probability(\rvbzo = \svbzo | \rvt = t, \rvbb = \svbb, \rvbzi = \svbzi)\\
    & \sequal{(d)} \Probability(\rvbz = \svbz | \rvt = t, \rvbb = \svbb), \label{eq_z_split_indep_parts}
\end{align}
where $(a)$ and $(d)$ follow from the definition of conditional probability, $(b)$ follows from \cref{item:rule_2}, \cref{eq_condition}, and \cref{fact}, and $(c)$ follows from \cref{item:rule_2} and \cref{lemma_supp_dsep}\cref{item:4}. Likewise, we simplify the second term in \cref{eq_z_t_marginalized} as follows:
\begin{align}
    \Probability(\rvbb = \svbb| \doot) \sequal{(a)} \Probability(\rvbb = \svbb| \rvt = t), \label{eq_b_dot}
\end{align}
where $(a)$ follows from \cref{item:rule_2} and \cref{lemma_supp_dsep}\cref{item:3}.\\

\noindent Putting together \cref{eq_z_t_marginalized,eq_z_split_indep_parts,eq_b_dot}, results in \cref{eq_second_term} as follows:
\begin{align}
    \Probability(\rvbz = \svbz | \doot) & = \sum_{\svbb} \Probability(\rvbz = \svbz | \rvt = t, \rvbb = \svbb) \Probability(\rvbb = \svbb| \rvt = t) \sequal{(a)} \Probability(\rvbz = \svbz | \rvt = t),
\end{align}
where $(a)$ follows from the law of total probability.

\subsection{Proof of \cref{eq_thm_main_2}}
 First, using the law of total probability, we have
\begin{align}
    \Probability(\rvy = y |\doot) & =  \sum_{\svbb} \Probability(\rvy = y | \doot, \rvbb = \svbb) \Probability(\rvbb = \svbb| \doot). \label{eq_y_t_marginalized_via_b}
\end{align}
Now, we show that the first term in RHS of \cref{eq_y_t_marginalized_via_b} can be simplified as follows
\begin{align}
    & \Probability(\rvy = y | \doot, \rvbb = \svbb) \\
    & = \sum_{\svbzi} \Bigparenth{\sum_{t'} \Probability(\rvy = y |\rvbs = \svbs, \rvt = t') \Probability(\rvt = t')} \Probability(\rvbzi = \svbzi | \rvbb = \svbb, \rvt = t). \label{eq_third_term}
\end{align}
where $\rvbs \defn (\rvbb, \rvbzi)$. Using \cref{eq_b_dot} and \cref{eq_third_term} in \cref{eq_y_t_marginalized_via_b}, completes the proof of \cref{eq_thm_main_2} as follows:
\begin{align}
     & \Probability(\rvy = y |\doot) \\
     & = \sum_{\svbs} \Bigparenth{\sum_{t'} \Probability(\rvy = y |\rvbs = \svbs, \rvt = t') \Probability(\rvt = t')} \Probability(\rvbzi = \svbzi | \rvbb = \svbb, \rvt = t) \Probability(\rvbb = \svbb| \rvt = t)\\
     & \sequal{(a)} \sum_{\svbs} \Bigparenth{\sum_{t'} \Probability(\rvy = y | \rvbs = \svbs, \rvt = t') \Probability(\rvt = t')} \Probability(\rvbs = \svbs | \rvt = t),
\end{align}
where $(a)$ follows from the definition of conditional probability.

\paragraph{Proof of \cref{eq_third_term}:}
From the law of total probability, we have
\begin{align}
    & \Probability(\rvy = y | \doot, \rvbb = \svbb) \\
    & \!=\! \sum_{\svbzi} \Probability(\rvy = y | \rvbb = \svbb, \rvbzi \!=\! \svbzi, \doot) \Probability(\rvbzi \!=\! \svbzi| \rvbb = \svbb, \doot). \label{eq_y_b_dot_marginalized}
\end{align}
Now, we simplify the first term in \cref{eq_y_b_dot_marginalized} as follows:
\begin{align}
   & \Probability(\rvy = y | \rvbb = \svbb, \rvbzi = \svbzi, \doot) \\
   & \sequal{(a)} \Probability(\rvy = y | \doob, \rvbzi = \svbzi, \doot) \\
   & \sequal{(b)} \Probability(\rvy = y | \doob, \rvbzi = \svbzi) \\
   & \sequal{(c)} \sum_{t'} \Probability(\rvy = y | \doob, \rvbzi = \svbzi, \rvt = t') \Probability(\rvt = t' | \doob, \rvbzi = \svbzi), \label{eq_fourth_term}
\end{align}
where $(a)$ follows from \cref{item:rule_2}, \cref{lemma_supp_dsep}\cref{item:6}, and \cref{fact}, $(b)$ follows from \cref{item:rule_3a} and \cref{lemma_supp_dsep}\cref{item:1}, and $(c)$ follows from the law of total probability. We further simplify the first term in \cref{eq_fourth_term} as follows:
\begin{align}
    \Probability(\rvy = y | \doob, \rvbzi = \svbzi, \rvt = t') & \sequal{(a)} \Probability(\rvy = y | \rvbb = \svbb, \rvbzi = \svbzi, \rvt = t'), \label{eq_y_z_t_dob}
\end{align}
where $(a)$ follows from \cref{item:rule_2} and \cref{lemma_supp_dsep}\cref{item:6}. Using \cref{eq_y_z_t_dob} and \cref{eq_y_z_marginalized_1} in  \cref{eq_fourth_term}, we have
\begin{align}
   \Probability(\rvy = y | \rvbb = \svbb, \rvbzi = \svbzi, \doot) = \sum_{t'} \Probability(\rvy = y | \rvbb = \svbb, \rvbzi = \svbzi, \rvt = t') \Probability(\rvt = t'). \label{eq_fourth_term_simplified}
\end{align}
Now, we simplify the second term in \cref{eq_y_b_dot_marginalized} as follows:
\begin{align}
   \Probability(\rvbzi  = \svbzi| \rvbb = \svbb, \doot) \sequal{(a)} \Probability(\rvbzi  = \svbzi| \rvbb = \svbb,  \rvt = t), \label{eq_zi_dot_b}
\end{align}
where $(a)$ follows from \cref{item:rule_2} and \cref{lemma_supp_dsep}\cref{item:4}.\\

\noindent Putting together \cref{eq_y_b_dot_marginalized,eq_fourth_term_simplified,eq_zi_dot_b} results in \cref{eq_third_term}.

\subsection{Necessity of Assumption \ref{assumption_confounded}}
\label{app_ass_confounded}
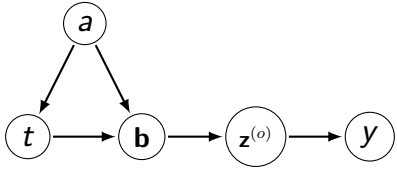
\begin{figure}[h]
    \centering
    \begin{tikzpicture}[scale=0.75, every node/.style={transform shape}, > = latex, shorten > = 1pt, shorten < = 1pt]
	\node[shape=circle,draw=black](y) at (6,-0.5) {\LARGE$\rvy$};
	\node[shape=circle,draw=black](ztilde) at (4,-0.5) {\large$\rvbzo$};
	\node[shape=circle,draw=black](a) at (1.0,1.5) {\LARGE$\rva$};
	\node[shape=circle,draw=black](b) at (2,-0.5) {\Large$\rvbb$};
	\node[shape=circle,draw=black](t) at (0,-0.5) {\LARGE$\rvt$};
	\path[style=thick][<->, bend right, white](t) [dashed] edge (y);
	\path[style=thick][->](t) edge (b);
	\path[style=thick][->](b) edge (ztilde);
	\path[style=thick][->](ztilde) edge (y);
	\path[style=thick][<-](b) edge (a);
	\path[style=thick][->](a) edge (t);
	\end{tikzpicture}  
    \caption{An SMCM signifying the importance of Assumption \ref{assumption_confounded}}
    \label{fig:assumption_importance}
\end{figure}

In this section, we provide an example to signify the importance of Assumption \ref{assumption_confounded} to Theorem \ref{thm_main}. Consider the semi-Markovian causal model in Figure \ref{fig:assumption_importance} where Assumptions \ref{assumption_descendant} and \ref{assumption_known_children} hold but Assumption \ref{assumption_confounded} does not hold. 
\\
While $\rvbz = (\rvbzi, \rvbzo)$ satisfies \cref{eq_ci,eq_condition} where $\rvbzi = \emptyset$, the causal effect is not equal to the formulae in \cref{eq_thm_main_1} or \cref{eq_thm_main_2}. To see this, we note that the set $\{ \rva \}$ is a back-door set in Figure \ref{fig:assumption_importance} implying
\begin{align}
    \Probability(\rvy | \doot) = \sum_{a}\Probability(\rvy | \rva, \rvt) \Probability(\rva). \label{eq_bd_example}
\end{align}
Now, we simplify the right hand side of \cref{eq_bd_example} to show explicitly that it is not equivalent to \cref{eq_thm_main_1}. From the law of total probability, we have
\begin{align}
    \Probability(\rvy | \rva, \rvt) & = \sum_{\rvbz} \Probability(\rvy | \rvbz, \rva, \rvt) \Probability(\rvbz | \rva, \rvt)\\
    & \sequal{(a)} \sum_{\rvbz} \Bigparenth{\sum_{t'}  \Probability(\rvy | \rvbz, \rva, \rvt) \Probability(t')} \Probability(\rvbz | \rva, \rvt) \sequal{(b)} \sum_{\rvbz} \Bigparenth{\sum_{t'}  \Probability(\rvy | \rvbz, \rvt') \Probability(t')} \Probability(\rvbz | \rva, \rvt), \label{eq_back_door_simple_1}
\end{align}
where $(a)$ follows because $\sum_{t'} \Probability(t') = 1$ and $(b)$ follows $\rvy$ is independent of every other variable conditioned on $\rvbz$. Plugging \cref{eq_back_door_simple_1} in \cref{eq_bd_example}, we have 
\begin{align}
    \Probability(\rvy | \doot) = \sum_{\rvbz} \Bigparenth{\sum_{t'}  \Probability(\rvy | \rvbz, \rvt') \Probability(t')} \Bigparenth{\sum_{a}\Probability(\rvbz | \rva, \rvt) \Probability(\rva)}. \label{eq_bd_example_final}
\end{align}
Lastly, using the law of total probability, \cref{eq_thm_main_1} can be rewritten as 
\begin{align}
    \Probability(\rvy | \doot) = \sum_{\rvbz} \Bigparenth{\sum_{t'}  \Probability(\rvy | \rvbz, \rvt') \Probability(t')} \Bigparenth{\sum_{a}\Probability(\rvbz | \rva, \rvt) \Probability(\rva | \rvt)}. \label{eq_bd_example_alternative}
\end{align}
Therefore, the variables $\rva$ and $\rvt$ could be such that \cref{eq_bd_example_final} is different from \cref{eq_bd_example_alternative}. We note that similar steps can be used to show that \cref{eq_bd_example} is not equivalent to \cref{eq_thm_main_2}. In conclusion, Assumption \ref{assumption_confounded} is crucial for the formulae in \cref{eq_thm_main_1,eq_thm_main_2} to hold.

\section{Proof of \cref{lemma_supp_dsep}}
\label{sec_proof_lemma_supp_dsep}
First, we state the following d-separation criterion used to prove \cref{lemma_supp_dsep}\cref{item:3} and \cref{lemma_supp_dsep}\cref{item:5}. See \cref{subsec_proof_extra_dsep} for a proof.
\begin{lemma}\label{lemma_extra_dsep}
Suppose \cref{assumption_descendant,assumption_confounded,assumption_known_children} hold. Then, $\rvt \dsep \rvbb | \rvbzi$ in $\cG_{\urvt}$.
\end{lemma}
\noindent Now, we prove each part of \cref{lemma_supp_dsep} one-by-one.
\paragraph{Proof of \cref{lemma_supp_dsep}\cref{item:1}} In $\cG^{\rvt}_{\brvbb}$, all edges going into $\rvbb$ are removed. Under \cref{assumption_known_children}, this implies that all edges going out of $\rvt$ are removed. Now, consider any path $\cP(F_{\rvt}, \rvy)$ between $F_{\rvt}$ and $\rvy$ in  $\cG^{\rvt}_{\brvbb}$. This path takes one of the following two forms: $(a)$ $F_{\rvt} \myrightarrow \rvt \myleftarrow \cdots \rvy$ or $(b)$ $F_{\rvt} \myrightarrow \rvt \doublearrow \cdots \rvy$. In either case, there is a collider at $\rvt$ in $\cP(F_{\rvt}, \rvy)$. This collider is blocked when $\rvbz$ and $\rvbb$ are conditioned on because $\rvt \notin \rvbz$, $\rvt \notin \rvbb$,  and $\rvt$ does not have any descendants in $\cG^{\rvt}_{\brvbb}$. Therefore, $\rvy \dsep F_{\rvt} | \rvbz, \rvbb$ in $\cG^{\rvt}_{\brvbb}$. Similarly, the collider is blocked when $\rvbzi$ and $\rvbb$ are conditioned on because $\rvt \notin \rvbzi$, $\rvt \notin \rvbb$,  and $\rvt$ does not have any descendants in $\cG^{\rvt}_{\brvbb}$. Therefore, $\rvy \dsep F_{\rvt} | \rvbzi, \rvbb$ in $\cG^{\rvt}_{\brvbb}$.

\paragraph{Proof of \cref{lemma_supp_dsep}\cref{item:3}} We prove this by contradiction. Assume there exists at least one unblocked path between $\rvt$ and some $\rvb \in \rvbb$ in $\cG_{\urvt}$. Let $\cP(\rvt, \rvb)$ denote any such unblocked path. 

Suppose we condition on $\rvbzi$. From \cref{lemma_extra_dsep}, $\cP(\rvt, \rvb)$ is blocked in $\cG_{\urvt}$ when $\rvbzi$ is conditioned on. Let $\rvv$ be any node at which $\cP(\rvt, \rvb)$ is blocked in $\cG_{\urvt}$ when $\rvbzi$ is conditioned on. We must have that $\rvv \in \cP(\rvt, \rvb) \setminus\{\rvt, \rvb\}$ and $\rvv \in \rvbzi$. Then, the path $\cP(\rvt, \rvv) \subset \cP(\rvt, \rvb)$ is unblocked in $\cG_{\urvt}$ when $\rvbzi$ is unconditioned on. However, this contradicts $\rvt \dsep \rvbzi$ in $\cG_{\urvt}$ (which follows from \cref{eq_condition}$(i)$ and \cref{fact}).

\paragraph{Proof of \cref{lemma_supp_dsep}\cref{item:4}} We prove this by contradiction. Assume there exists at least one unblocked path between $\rvt$ and some $\rvzi \in \rvbzi$ in $\cG_{\urvt}$ when $\rvbb$ is conditioned on. Let $\cP(\rvt, \rvzi)$ denote any such unblocked path. 

Suppose, we uncondition on $\rvbb$. From \cref{eq_condition}(i) and \cref{fact}, we have $\rvt \dsep \rvbzi$ in $\cG_{\urvt}$. Therefore, $\cP(\rvt, \rvzi)$ is blocked in $\cG_{\urvt}$ when $\rvbb$ is unconditioned on. Now, we create a set $\rvbv$ consisting of all the nodes at which $\cP(\rvt, \rvzi)$ is blocked in $\cG_{\urvt}$ when $\rvbb$ is unconditioned on. Define the set $\rvbv$ such that for any $\rvv \in \rvbv$, the following are true: $(a)$ $\rvv \in \cP(\rvt, \rvzi) \setminus\{\rvt, \rvzi\}$, $(b)$ $\cP(\rvt, \rvzi)$ contains a collider at $\rvv$ in $\cG_{\urvt}$, and $(c)$ there exists an unblocked 
descendant path from $\rvv$ to some $\rvb \in \rvbb$ in $\cG_{\urvt}$. 

Now, we must have $\rvbv \neq \emptyset$, since $\cP(\rvt, \rvzi)$ is blocked in $\cG_{\urvt}$ when $\rvbb$ is unconditioned on. Let $\rvv_c \in \rvbv$ be that node which is closest to $\rvt$ in the path  $\cP(\rvt, \rvzi)$, and let $\cP(\rvv_c, \rvb)$ be an unblocked descendant path from $\rvv$ to some $\rvb \in \rvbb$ in $\cG_{\urvt}$ (there must be one from the definition of the set $\rvbv$). Consider the path $\cP(\rvt, \rvb)$ obtained after concatenating $\cP(\rvt, \rvv_c)  \subset \cP(\rvt, \rvzi)$ and $\cP(\rvv_c, \rvb)$. By the definition of $\rvbv$ and the choice of $\rvv_c$, $\cP(\rvt, \rvb)$ is unblocked in $\cG_{\urvt}$ since $(a)$ $\cP(\rvt, \rvv_c)$ is unblocked in $\cG_{\urvt}$, $(b)$ $\cP(\rvv_c, \rvb)$ is unblocked in $\cG_{\urvt}$, and $(c)$ there is no collider at $\rvv_c$ in $\cP(\rvt, \rvb)$. However, this contradicts $\rvt \dsep \rvbb$ in $\cG_{\urvt}$ (which follows from \cref{lemma_supp_dsep}\cref{item:3}).

\paragraph{Proof of \cref{lemma_supp_dsep}\cref{item:5}} We prove this by contradiction. Assume there exists at least one unblocked path between $\rvt$ and $F_{\rvbb}$ in $\cG^{\rvbb}$ when $\rvbzi$ is conditioned on. Let $\cP(\rvt, F_{\rvbb})$ denote the shortest of these unblocked path. By definition of $\cG^{\rvbb}$, this path has to be of the form: $\rvt \cdots, \rvb \myleftarrow F_{\rvbb}$ for some $\rvb \in \rvbb$. Now, we have the following three cases: 
\begin{enumerate}
    \item[(i)] $\cP(\rvt, F_{\rvbb})$ contains $\rvt \myrightarrow \rvb$: In this case, because a path is a sequence of distinct nodes, $\cP(\rvt, F_{\rvbb})$ has to be $\rvt \myrightarrow \rvb \myleftarrow F_{\rvbb}$. By assumption,  $\cP(\rvt, F_{\rvbb})$ is unblocked when $\rvbzi$ is conditioned on. Since there is a collider at $\rvb$ in $\cP(\rvt, F_{\rvbb})$, there exists at least one unblocked descendant path from $\rvb$ to $\rvbzi$ when $\rvbzi$ is conditioned on. Let $\cP(\rvb, \rvzi)$ denote the shortest of these paths from $\rvb$ to some $\rvzi \in \rvbzi$ in $\cG^{\rvbb}$. We note that this path also exists in $\cG$ and is of the form $\rvb \myrightarrow \cdots \myrightarrow \rvzi$
    
    Suppose we uncondition on $\rvbzi$. Consider the path $\cP(\rvt, \rvzi) \supset \cP(\rvb, \rvzi)$ between $\rvt$ and $\rvzi$ of the form $\rvt \myrightarrow \rvb \myrightarrow \cdots \myrightarrow \rvzi$ in $\cG$. This path remains unblocked even when $\rvbzi$ is unconditioned on as it  does not have any colliders. This contradicts $\rvzi \dsep \rvt$ (which follows from \cref{eq_condition}).
    
    \item[(ii)] $\cP(\rvt, F_{\rvbb})$ contains $\rvt \myrightarrow \rvb_1$ for some $\rvb_1 \in \rvbb$ such that $\rvb_1 \neq \rvb$: In this case, the path $\cP(\rvt, F_{\rvbb})$ has to be of the form $\rvt \myrightarrow \rvb_1 \cdots \rvb \myleftarrow F_{\rvbb}$. Therefore, there exists at least one collider on the path $\cP(\rvt, F_{\rvbb})$. Let $\rvv \in \cP(\rvt, F_{\rvbb}) \setminus\{\rvt, F_{\rvbb}\}$ be the collider on the path $\cP(\rvt, F_{\rvbb})$ that is closest to $\rvb_1$. 
    Consider the path $\cP(\rvt, \rvv) \subset \cP(\rvt, F_{\rvbb})$. We note that this path also exists in $\cG$ and is of the form $\rvt \myrightarrow \rvb_1 \myrightarrow \cdots \myrightarrow \rvv$.
    
    By assumption,  $\cP(\rvt, F_{\rvbb})$ is unblocked when $\rvbzi$ is conditioned on. Since there is a collider at $\rvv$ in $\cP(\rvt, F_{\rvbb})$, there exists at least one unblocked descendant path from $\rvv$ to $\rvbzi$ when $\rvbzi$ is conditioned on. Let $\cP(\rvv, \rvzi)$ denote the shortest of these paths from $\rvv$ to some $\rvzi \in \rvbzi$ in $\cG^{\rvbb}$. We note that this path also exists in $\cG$ and is of the form $\rvv \myrightarrow \cdots \myrightarrow \rvzi$.
    
    Suppose we uncondition on $\rvbzi$. Consider the path $\cP(\rvt, \rvzi)$ between $\rvt$ and $\rvzi$ in $\cG$ obtained after concatenating $\cP(\rvt, \rvv)  \subset \cP(\rvt, F_{\rvbb})$ and $\cP(\rvv, \rvzi)$. This path, of the form $\rvt \myrightarrow \rvb_1 \myrightarrow \cdots \myrightarrow \rvv \myrightarrow \cdots \myrightarrow \rvzi$, remains unblocked even when $\rvbzi$ is unconditioned on as it  does not have any colliders. This contradicts $\rvzi \dsep \rvt$ (which follows from \cref{eq_condition}).
    
    \item[(ii)] $\cP(\rvt, F_{\rvbb})$ does not contain $\rvt \myrightarrow \rvb_1$ for every $\rvb_1 \in \rvbb$: By assumption,  $\cP(\rvt, F_{\rvbb})$ is unblocked in $\cG^{\rvbb}$ when $\rvbzi$ is conditioned on. Therefore, if $\cP(\rvt, F_{\rvbb})$ does not contain the edge $\rvt \myrightarrow \rvb_1$ for any $\rvb_1 \in \rvbb$, there exists a path $\cP(\rvt, \rvb)$ between $\rvt$ to $\rvb$ in $\cG$ that is unblocked when $\rvbzi$ is conditioned on, and takes one of the following two forms: $(a)$ $\rvt \myleftarrow \cdots \rvb$ or $(b)$ $\rvt \doublearrow \cdots \rvb$. Then, it is easy to see that the path $\cP(\rvt, \rvb)$ also remains unblocked in $\cG_{\urvt}$ while $\rvbzi$ is conditioned on. However, this contradicts $\rvt \dsep \rvbb | \rvbzi$ in $\cG_{\urvt}$ (which follows from \cref{lemma_extra_dsep}).
\end{enumerate}

\paragraph{Proof of \cref{lemma_supp_dsep}\cref{item:6}} We prove this by contradiction. Assume there exists at least one unblocked path between $\rvy$ and some $\rvb \in \rvbb$ in $\cG_{\urvbb}$ when $\rvt$ and $\rvbzi$ are conditioned on. Let $\cP(\rvb, \rvy)$ denote the shortest of these unblocked path. Therefore, no $\rvb_1 \in \rvbb$, such that $\rvb_1 \neq \rvb$, is on the path $\cP(\rvb, \rvy)$, i.e., $\rvb_1 \notin \cP(\rvb, \rvy)$. Further, $\cP(\rvb, \rvy)$ takes one of the following two forms because all the edges going out of $\rvbb$ are removed in $\cG_{\urvbb}$: $(a)$ $\rvb \myleftarrow \cdots \rvy$ or $(b)$ $\rvb \doublearrow \cdots \rvy$.

Suppose we condition on $\rvbzo$ (while $\rvt$ and $\rvbzi$ are still conditioned on). From \cref{eq_ci} and \cref{fact}, we have $\rvy \dsep \rvb | \rvt, \rvbz$ in $\cG_{\urvbb}$. Therefore, the path  $\cP(\rvb, \rvy)$ is blocked in $\cG_{\urvbb}$ when $\rvbzo$ is conditioned on (while $\rvt$ and $\rvbzi$ are still conditioned on). Let $\rvv$ be any node at which $\cP(\rvb, \rvy)$ is blocked in $\cG_{\urvbb}$ when $\rvbzo$ is conditioned on (while $\rvt$ and $\rvbzi$ are still conditioned on). We must have that $\rvv \in \cP(\rvb, \rvy) \setminus\{\rvy, \rvb\}$ and $\rvv \in \rvbzo$. Suppose we uncondition on $\rvbzo$ (while $\rvt$ and $\rvbzi$ are still conditioned on). Then, the path $\cP(\rvb, \rvv) \subset \cP(\rvb, \rvy)$ is unblocked in $\cG_{\urvbb}$. 

We consider the following two scenarios depending on whether or not $\cP(\rvb, \rvv)$ contains $\rvt$. In both scenarios, we show that there is an unblocked path between $\rvt$ and $\rvv$ in $\cG_{\urvbb}$ when we condition on $\rvbb$ (while $\rvt$ and $\rvbzi$ are still conditioned on).

\begin{enumerate}
    \item[(i)]  $\cP(\rvb, \rvv)$ contains $\rvt$: Consider the path $\cP(\rvt, \rvv) \subset \cP(\rvb, \rvv)$ which is unblocked in $\cG_{\urvbb}$ when $\rvt$ and $\rvbzi$ are conditioned on. Further, by the choice of $\cP(\rvb, \rvy)$, no $\rvb_1 \in \rvbb$ is on the path $\cP(\rvt, \rvv)$. Therefore, the path $\cP(\rvt, \rvv)$ in $\cG_{\urvbb}$ remains unblocked when we condition on $\rvbb$ (while $\rvt$ and $\rvbzi$ are still conditioned on). 
    
    \item[(ii)]  $\cP(\rvb, \rvv)$ does not contain $\rvt$: Consider the path $\cP(\rvt, \rvv) \supset \cP(\rvb, \rvv)$ (by including the extra edge $\rvt \rightarrow \rvb$) which takes one of the following two forms: $(a)$ $\rvt \myrightarrow \rvb \myleftarrow \cdots \rvv$ or $(b)$ $\rvt \myrightarrow \rvb \doublearrow \cdots \rvv$. Further, by the choice of $\cP(\rvb, \rvy)$, no $\rvb_1 \in \rvbb$ ($\rvb_1 \neq \rvb$) is on the path $\cP(\rvt, \rvv)$. Suppose we condition on $\rvbb$ (while $\rvt$ and $\rvbzi$ are still conditioned on). Then, the path $\cP(\rvt, \rvv)$  in $\cG_{\urvbb}$ is unblocked because $(a)$ the collider at $\rvb$ is unblocked when $\rvbb$ is  conditioned on and $(b)$ the path $\cP(\rvb, \rvv)$  in $\cG_{\urvbb}$ remains unblocked when $\rvbb$ is  conditioned on (while $\rvt$ and $\rvbzi$ are still conditioned on).
\end{enumerate}   

Now, suppose we uncondition on $\rvt$ (while $\rvbb$ and $\rvbzi$ are still conditioned on). We have the following two scenarios depending on whether or not $\cP(\rvt, \rvv)$ in $\cG_{\urvbb}$ remains unblocked. In both scenarios, we show that there is an unblocked path between $\rvt$ and $\rvv$ in $\cG_{\urvbb}$ when we uncondition on $\rvt$ (while $\rvbb$ and $\rvbzi$ are still conditioned on).
    \begin{enumerate}
        \item If $\cP(\rvt, \rvv)$ remains unblocked: In this case,  $\cP(\rvt, \rvv)$ in $\cG_{\urvbb}$ is an unblocked path between $\rvt$ and $\rvv$ when $\rvbzi$ and $\rvbb$ are conditioned on, as desired.
        \item If $\cP(\rvt, \rvv)$ does not remain unblocked: In this case, it is the unconditioning on $\rvt$ (while $\rvbb$ and $\rvbzi$ are still conditioned on) that blocks $\cP(\rvt, \rvv)$. Now, we create a set $\rvbw$ consisting of all the nodes at which $\cP(\rvt, \rvv)$ is blocked in $\cG_{\urvbb}$ when $\rvt$ is unconditioned on (while $\rvbb$ and $\rvbzi$ are still conditioned on). Define the set $\rvbw$ such that for any $\rvw \in \rvbw$, the following are true: $(a)$ $\rvw \in \cP(\rvt, \rvv) \setminus\{\rvt, \rvv\}$, $(b)$ $\cP(\rvt, \rvv)$ contains a collider at $\rvw$ in $\cG_{\urvbb}$, and $(c)$ there exists an unblocked  descendant path from $\rvw$ to $\rvt$ in $\cG_{\urvbb}$. 
        
        Now, we must have $\rvbw \neq \emptyset$, since $\cP(\rvt, \rvv)$ is blocked in $\cG_{\urvbb}$ when $\rvt$ is unconditioned on (while $\rvbb$ and $\rvbzi$ are still conditioned on). Let $\rvw_c \in \rvbw$ be that node which is closest to $\rvv$ in the path  $\cP(\rvt, \rvv)$, and let $\cP(\rvw_c, \rvt)$ be an unblocked descendant path from $\rvw_c$ to $\rvt$ in $\cG_{\urvbb}$ (there must be one from the definition of the set $\rvbw$). Consider the path $\cP'(\rvv, \rvt)$ obtained after concatenating $\cP(\rvv, \rvw_c)  \subset \cP(\rvt, \rvv)$ and $\cP(\rvw_c, \rvt)$. By the definition of $\rvbw$ and the choice of $\rvw_c$, $\cP'(\rvv, \rvt)$ is unblocked in $\cG_{\urvbb}$ when $\rvt$ is unconditioned on (while $\rvbb$ and $\rvbzi$ are still conditioned on) since $(a)$ $\cP(\rvv, \rvw_c)$ is unblocked, $(b)$ $\cP(\rvw_c, \rvt)$ is unblocked, and $(c)$ there is no collider at $\rvw_c$ in $\cP'(\rvv, \rvt)$. 
        Therefore, we have an unblocked path between $\rvt$ and $\rvv$ in $\cG_{\urvbb}$ when $\rvbzi$ and $\rvbb$ are conditioned on, as desired.
    \end{enumerate}
    
    To conclude the proof, we note that the existence of an unblocked path between $\rvt$ and $\rvv \in \rvbzo$ in $\cG_{\urvbb}$ when $\rvbzi$ and $\rvbb$ are conditioned on contradicts $\rvbzo \dsep \rvt | \rvbb, \rvbzi$ in $\cG_{\urvbb}$ (which follows from \cref{eq_condition} and \cref{fact}).
    
\subsection{Proof of \cref{lemma_extra_dsep}}
\label{subsec_proof_extra_dsep}
First, we claim $\rvt \dsep \rvbb | \rvbz$ in $\cG_{\urvt}$. We assume this claim and proceed to prove the statement in the Lemma by contradiction. Assume there exists at least one unblocked path between $\rvt$ and some $\rvb \in \rvbb$ in $\cG_{\urvt}$ when $\rvbzi$ is conditioned on. Let $\cP(\rvt, \rvb)$ denote the shortest of these unblocked path. Therefore, no $\rvb_1 \in \rvbb$ such that $\rvb_1 \neq \rvb$ is not on the path  $\cP(\rvt, \rvb)$, i.e., $\rvb_1 \notin \cP(\rvt, \rvb)$.

Suppose we condition on $\rvbzo$ (while $\rvbzi$ is still conditioned on). From the claim, $\cP(\rvt, \rvb)$ is blocked in $\cG_{\urvt}$ when $\rvbzo$ is conditioned on (while $\rvbzi$ is still conditioned on). Let $\rvv$ be any node at which $\cP(\rvt, \rvb)$ is blocked in $\cG_{\urvt}$ when $\rvbzo$ is conditioned on (while $\rvbzi$ is still conditioned on). We must have that $\rvv \in \cP(\rvt, \rvb) \setminus\{\rvt, \rvb\}$ and $\rvv \in \rvbzo$. Then, the path $\cP(\rvt, \rvv) \subset \cP(\rvt, \rvb)$ is unblocked in $\cG_{\urvt}$ when $\rvbzo$ is unconditioned on (while $\rvbzi$ is still conditioned on).
Further, no $\rvb \in \rvbb$ is on the path  $\cP(\rvt, \rvv)$. As a result, the path $\cP(\rvt, \rvv)$ remains unblocked when $\rvbb$ is conditioned on (while $\rvbzi$ is still conditioned on). However, this contradicts $\rvt \dsep \rvbzo | \rvb, \rvbzi$ in $\cG_{\urvt}$ (which follows from \cref{eq_condition}$(ii)$ and \cref{fact}).\\

\noindent \textbf{Proof of Claim - $\rvt \dsep \rvbb | \rvbz$ in $\cG_{\urvt}$: } It remains to prove the claim $\rvt \dsep \rvbb | \rvbz$ in $\cG_{\urvt}$. We prove this by contradiction. Assume there exists at least one unblocked path between $\rvt$ and some $\rvb \in \rvbb$ in $\cG_{\urvt}$ when $\rvbz$ is conditioned on. Let $\cP(\rvt, \rvb)$ denote any such unblocked path. This path takes one of the following two forms: $(a)$ $\rvt \myleftarrow \cdots \rvb$ or $(b)$ $\rvt \doublearrow \cdots \rvb$ because all edges going out of $\rvt$ are removed in $\cG_{\urvt}$. 

Suppose we condition on $\rvt$ (while $\rvbz$ is still conditioned on).  The path $\cP(\rvt, \rvb)$ remains unblocked because $\rvt \notin \cP(\rvt, \rvb)$ (a path is a sequence of distinct nodes). Then, the path $\cP(\rvy, \rvb) \supset \cP(\rvt, \rvb)$ of the form $(a)$ $\rvy \doublearrow \rvt \myleftarrow \cdots \rvb$ or $(b)$ $\rvy \doublearrow \rvt \doublearrow \cdots \rvb$ is unblocked because the additional conditioning on $\rvt$ (while $\rvbz$ is still conditioned on) unblocks the collider at $\rvt$. However, this contradicts $\rvbb \dsep \rvy | \rvt, \rvbz$ in $\cG_{\urvt}$ (which follows from \cref{eq_ci} and \cref{fact}).

\section{Experimental Results}
\label{appendix_experiments}
In this section, we provide additional experimental results. First, we provide more details regarding the numerical example in Section \ref{subsubsec_causal_effect_Estimation_post_treatment}. Next, we demonstrate the applicability of our method on a class of graphs slightly different from the one in Section \ref{subsec_random_graphs}. Then, we provide the 6 random graphs from Section \ref{sec:descsynth} as well as ATE estimation results on  specific choices of SMCMs including the one in Figure \ref{fig_toy_example}. Finally, we provide histograms analogous to Figure \ref{fig_german_1} for the second choice of $\rvbb$ on German credit dataset as well as details about our analysis with Adult dataset. 
\subsection{Numerical example in Section \ref{subsubsec_causal_effect_Estimation_post_treatment}}
The observed variables for this example also follow the structural equation model in \eqref{syn:func_model}. Also, to generate the true ATE, we intervene on the generation model in \cref{syn:func_model} by setting $t=0$ and $t = 1$. 

\subsection{Applicability to a class of random graphs}
As in Section \ref{subsec_random_graphs}, we create a class of random SMCMs, sample 100 SMCMs from this class, and check if \cref{eq_ci,eq_condition} hold by checking for corresponding d-separations in the SMCMs. The class of random graphs considered here is analogous to the class of random graphs considered in Section \ref{subsec_random_graphs} expect for the choice of $\rvt$. Here, we choose any variable that is ancestor of $\rvy$ but not its parent as $\rvt$. This is in contrast to Section \ref{subsec_random_graphs} where we choose any variable that is ancestor of $\rvy$ but not its parent or grandparent as $\rvt$. We compare the success rate of the same two approaches: $(i)$ exhaustive search for $\rvbz$ satisfying \cref{eq_ci,eq_condition}  and $(ii)$ search for a $\rvbz$ of size at-most 5 satisfying \cref{eq_ci,eq_condition}. We provide the number of successes of these approaches as a tuple in Table \ref{table_success_rate_2} for various $p$, $d$, and $q$. As before, we see that the two approaches have comparable performances and the IDP algorithm gives 0 successes across various $p$, $d$, and $q$ even though it is supplied with the true PAG. Also, as expected the number of successes for this class of graphs is much lower than the class considered in Section \ref{subsec_random_graphs}.

\begin{table}[h]
  \caption{Number of successes out of 100 random graphs for our methods shown as a tuple. The first method searches a $\rvbz$ exhaustively and the second method searches a $\rvbz$ with size at-most 5.} 
  \label{table_success_rate_2}
  \vspace{1mm}
\centering
  \begin{tabular}{ccccccccc}
    \toprule
    \multirow{3}{*}{} &
      \multicolumn{3}{c}{$p = 10$} &
      \multicolumn{3}{c}{$p = 15$} \\
      &  $d = 2$ & $d = 3$ & $d = 4$ & $d = 2$ & $d = 3$ & $d = 4$  \\
      \midrule
    \midrule
    $q = 0.0$ & $(6,6)$ & $(3,3)$ & $(1, 1)$ & $(11, 11)$ & $(2, 2)$ & $(1, 1)$\\
    $q = 0.5$ & $(3, 3)$ & $(0, 0)$  & $(0, 0)$ & $(5, 5)$ & $(2, 2)$ & $(1, 1)$\\
    $q = 1.0$ & $(1, 1)$ & $(0, 0)$ & $(0, 0)$ & $(1, 1)$ & $(0, 0)$ & $(0, 0)$\\
    \bottomrule
  \end{tabular}
\end{table}

\subsection{ATE estimation}
\begin{figure}[h]
  \centering
  \begin{tabular}{ccc}
	\begin{tikzpicture}[scale=0.6, every node/.style={transform shape}, > = latex, shorten > = 1pt, shorten < = 1pt]
	\node[shape=circle,draw=black](x1) at (0,1.5) {\LARGE$\rvx_1$};
	\node[shape=circle,draw=black](x2) at (2,1.5) {\LARGE$\rvx_2$};
	\node[shape=circle,draw=black](y) at (6,-1.5) {\LARGE$\rvy$};
	\node[shape=circle,draw=black](ztilde) at (4,-1.5) {\Large$\rvbzo$};
	\node[shape=circle,draw=black](zhat) at (4,0.5) {\Large$\rvbzi$};
	\node[shape=circle,draw=black](b) at (2,-1.5) {\Large$\rvbb$};
	\node[shape=circle,draw=black](t) at (0,-1.5) {\LARGE$\rvt$};
	\path[style=thick][->](x1) edge (t);
	\path[style=thick][<->, bend right](t) [dashed] edge (y);
	\path[style=thick][<->, bend left](x1) [dashed] edge (x2);
	\path[style=thick][->, red](x1) edge (x2);
	\path[style=thick][->](t) edge (b);		
	\path[style=thick][<->](x2) [dashed] edge (b);		
	\path[style=thick][<->](t) [dashed] edge (x2);
	\path[style=thick][<->, bend left=20, red](zhat) [dashed] edge (y);
	\path[style=thick][->, red](zhat) edge (ztilde);
	\path[style=thick][->](b) edge (ztilde);
	\path[style=thick][->](ztilde) edge (y);
	\path[style=thick][<-](b) edge (zhat);
	\path[style=thick][->](zhat) edge (y);
	\path[style=thick][<->, bend left = 50](x2) [dashed] edge (y);
	\end{tikzpicture}&
    \begin{tikzpicture}[scale=0.6, every node/.style={transform shape}, > = latex, shorten > = 1pt, shorten < = 1pt]
	\node[shape=circle,draw=black](x1) at (0,1.5) {\LARGE$\rvx_1$};
	\node[shape=circle,draw=black](x2) at (2,1.5) {\LARGE$\rvx_2$};
	\node[shape=circle,draw=black](y) at (6,-1.5) {\LARGE$\rvy$};
	\node[shape=circle,draw=black](ztilde) at (4,-1.5) {\Large$\rvbzo$};
	\node[shape=circle,draw=black](zhat) at (4,0.5) {\Large$\rvbzi$};
	\node[shape=circle,draw=black](b) at (2,-1.5) {\Large$\rvbb$};
	\node[shape=circle,draw=black](t) at (0,-1.5) {\LARGE$\rvt$};
	\path[style=thick][->](x1) edge (t);
	\path[style=thick][<->, bend right](t) [dashed] edge (y);
	\path[style=thick][<->, bend left](x1) [dashed] edge (x2);
	\path[style=thick][->](t) edge (b);		
	\path[style=thick][<->, bend right=20, red](zhat) [dashed] edge (b);
	\path[style=thick][<->](x2) [dashed] edge (b);		
	\path[style=thick][<->](t) [dashed] edge (x2);
    \path[style=thick][->, red](zhat) edge (ztilde);
    \path[style=thick][->, red](zhat) edge (x2);
	\path[style=thick][->](b) edge (ztilde);
	\path[style=thick][->](ztilde) edge (y);
	\path[style=thick][<-](b) edge (zhat);
	\path[style=thick][->](zhat) edge (y);
	\path[style=thick][<->, bend left = 50](x2) [dashed] edge (y);
	\end{tikzpicture}&
	\begin{tikzpicture}[scale=0.6, every node/.style={transform shape}, > = latex, shorten > = 1pt, shorten < = 1pt]
	\node[shape=circle,draw=black](x1) at (0,1.5) {\LARGE$\rvx_1$};
	\node[shape=circle,draw=black](x2) at (2,1.5) {\LARGE$\rvx_2$};
	\node[shape=circle,draw=black](y) at (6,-1.5) {\LARGE$\rvy$};
	\node[shape=circle,draw=black](ztilde) at (4,-1.5) {\Large$\rvbzo$};
	\node[shape=circle,draw=black](zhat) at (4,0.5) {\Large$\rvbzi$};
	\node[shape=circle,draw=black](b) at (2,-1.5) {\Large$\rvbb$};
	\node[shape=circle,draw=black](t) at (0,-1.5) {\LARGE$\rvt$};
	\path[style=thick][->](x1) edge (t);
	\path[style=thick][->, bend left = 80, red](x1) edge (y);
    \path[style=thick][<->, bend right](t) [dashed] edge (y);
	\path[style=thick][<->, bend left](x1) [dashed] edge (x2);
	\path[style=thick][->](t) edge (b);		
	\path[style=thick][<->, bend right, red](zhat) [dashed] edge (ztilde);
	\path[style=thick][<->](x2) [dashed] edge (b);		
	\path[style=thick][<->](t) [dashed] edge (x2);
    \path[style=thick][->, red](zhat) edge (ztilde);
	\path[style=thick][->](b) edge (ztilde);
	\path[style=thick][->](ztilde) edge (y);
	\path[style=thick][<-](b) edge (zhat);
	\path[style=thick][->](zhat) edge (y);
	\path[style=thick][<->, bend left = 50](x2) [dashed] edge (y);
	\end{tikzpicture}
  \end{tabular}
  \caption{The causal graphs used to further validate our theoretical results. These are obtained by adding additional edges (shown in red) to $\cGtoy$ in \cref{fig_toy_example}. We denote these graphs (from left to right) by $\cGtoy_1$, $\cGtoy_2$, and $\cGtoy_3$, respectively.}
  \label{fig_more_examples}
\end{figure}
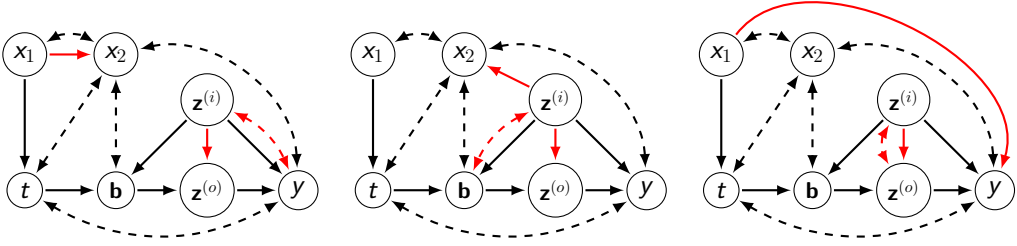

We also conduct ATE estimation experiments on four specific SMCMs. The first SMCM is the graph $\cG^{toy}$ in Figure \ref{fig_toy_example}. The remaining graphs, named $\cG^{toy}_i,~ i \in \{1,2,3\}$, are shown in Figure \ref{fig_more_examples}, and are obtained by adding additional edges and modifying $\cG^{toy}$. These SMCMs are designed in a way such that there exists $\rvbz = (\rvbzi,\rvbzo)$ satisfying the conditional independence statements in Theorem \ref{thm_main}.

\begin{figure}[h]
    \centering
    \begin{tabular}{cc}
    \includegraphics[width=0.45\linewidth,clip]{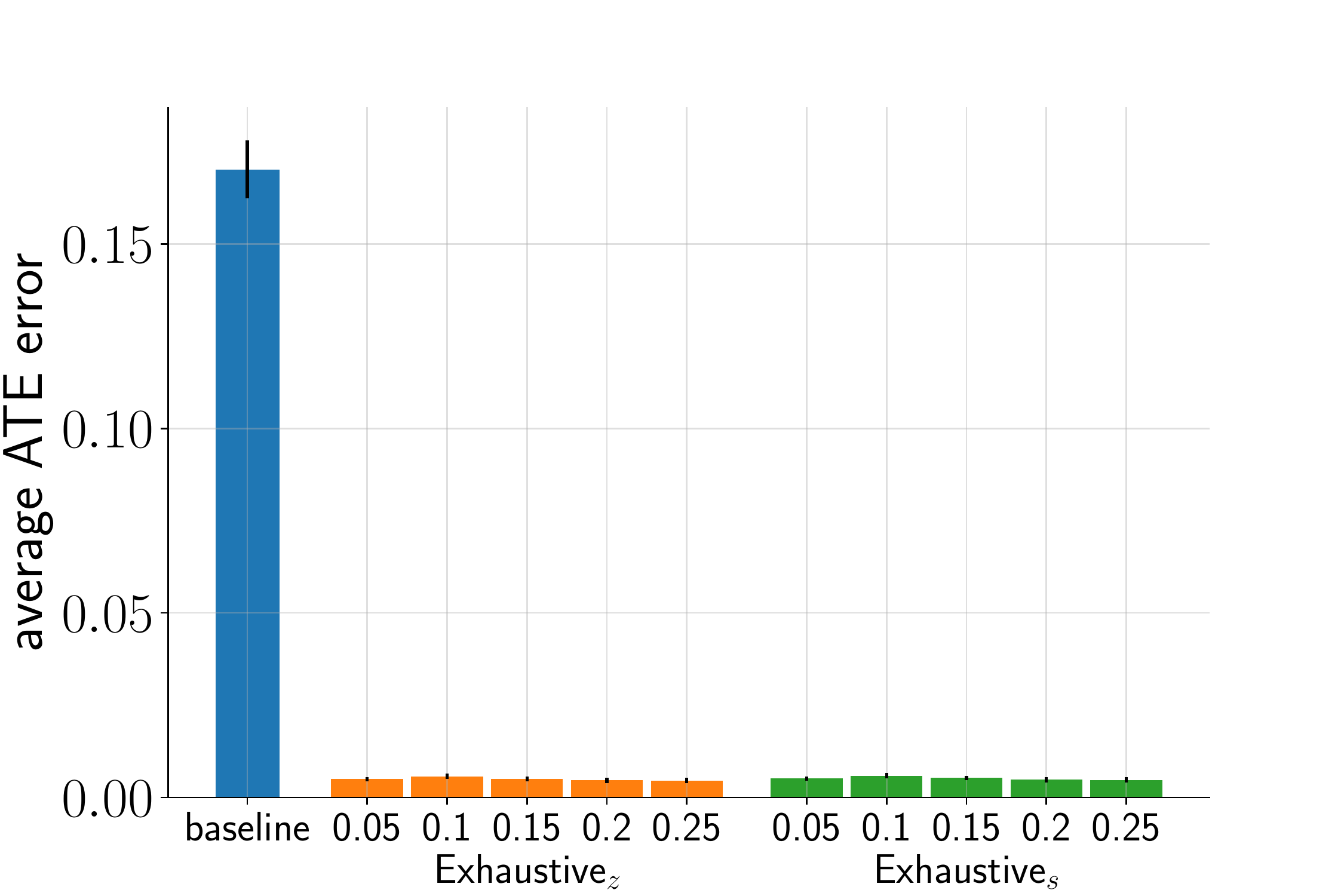}&
    \includegraphics[width=0.45\linewidth,clip]{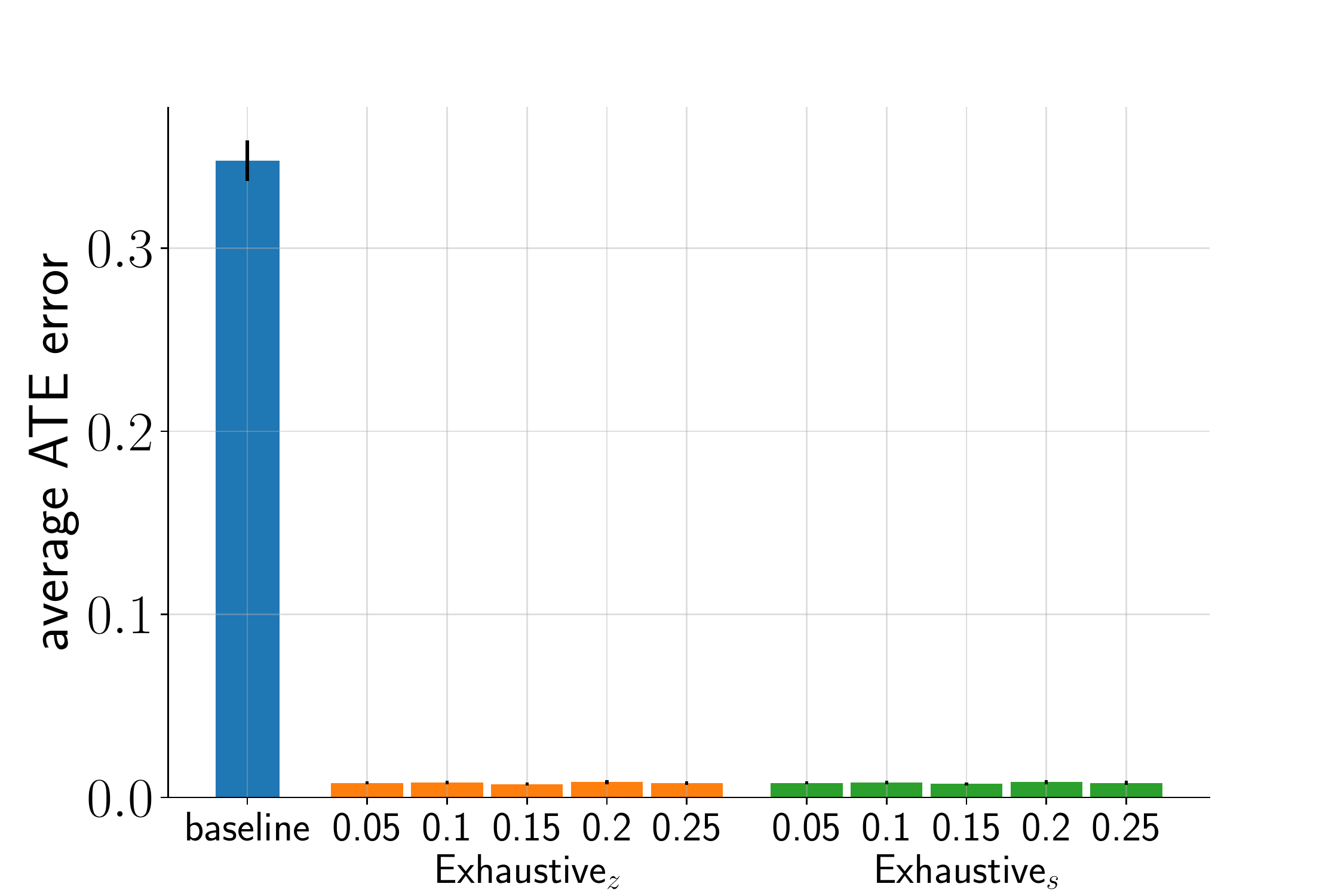}\\
    \includegraphics[width=0.45\linewidth,clip]{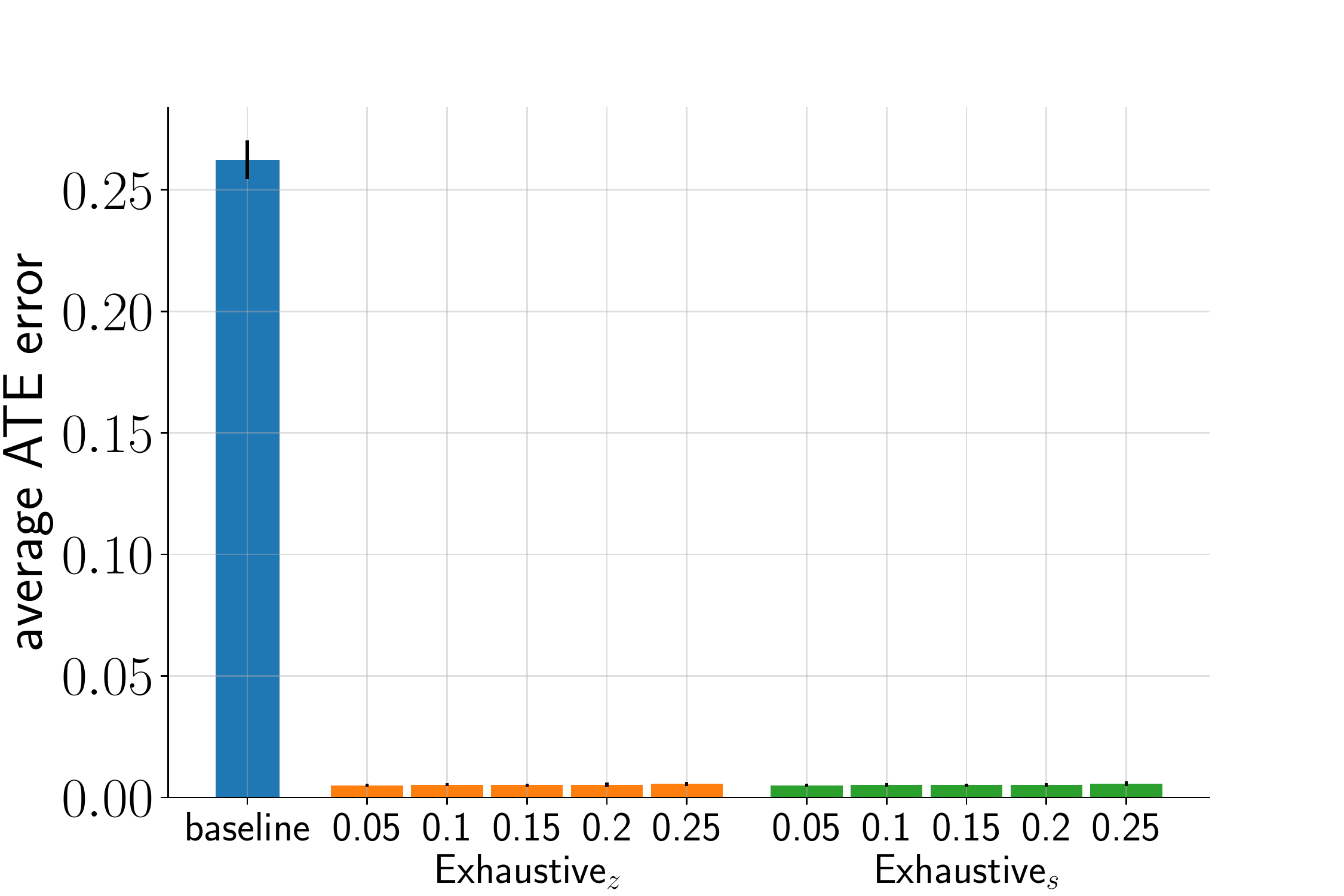}&
    \includegraphics[width=0.45\linewidth,clip]{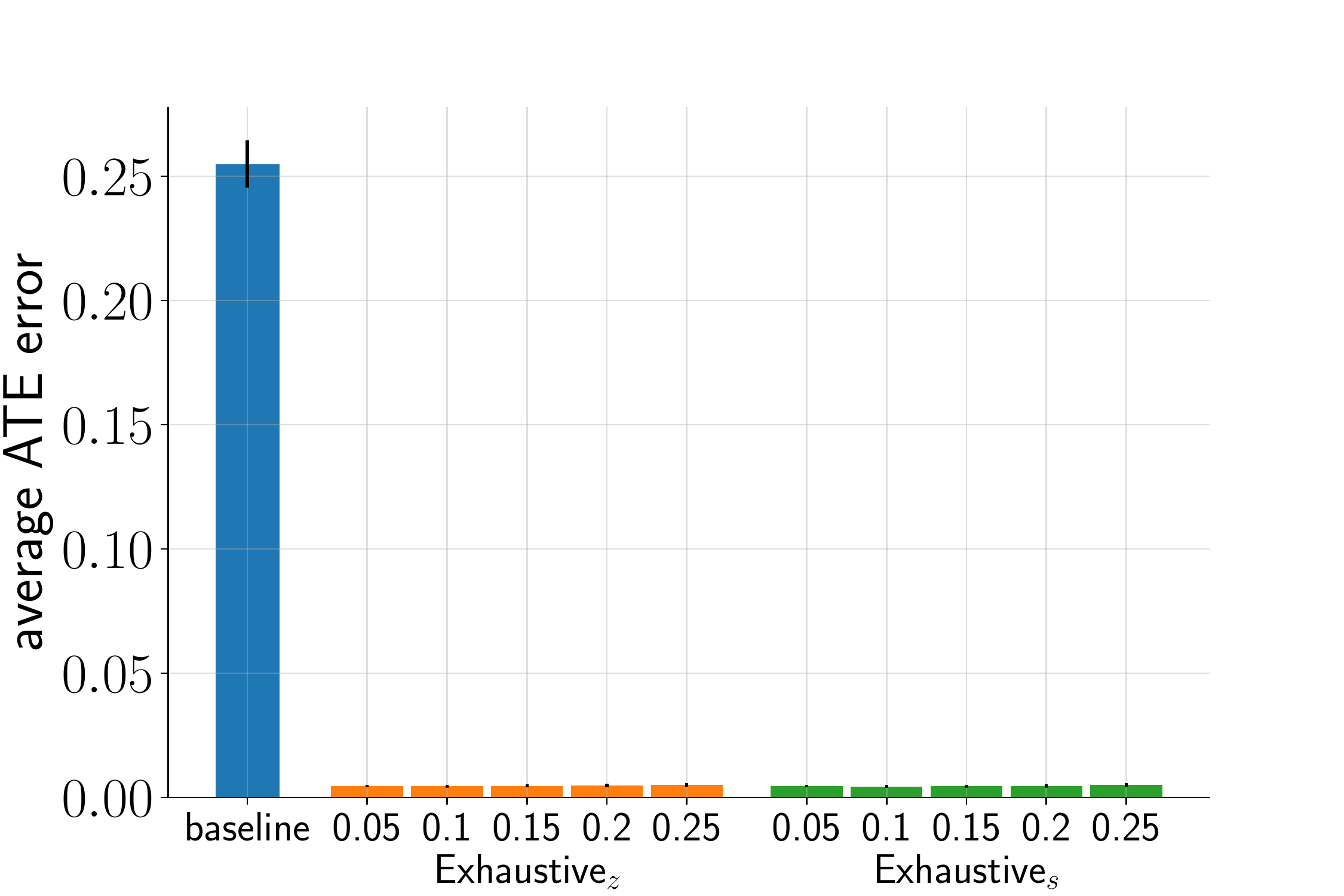}
    \end{tabular}
\caption{Performance of Algorithm \ref{alg:subset_search} for different p-value thresholds $p_v$ on $\cGtoy$ in \cref{fig_toy_example} on top left, on $\cGtoy_1$ from \cref{fig_more_examples} on top right, on $\cGtoy_2$ in \cref{fig_more_examples} on bottom left, and on $\cGtoy_3$ from \cref{fig_more_examples} on bottom right}
\label{fig_synthetic_1}
\end{figure}

We follow a data generation procedure similar to the one in Section \ref{sec:descsynth}. In contrast, we show the performance of our approach for a fixed $n$ but different thresholds of p-value $p_v$. We average the ATE error over 50 runs where in each run we set $n = 50000$. As we see in Figure \ref{fig_synthetic_1}, both the ATE estimates returned by Algorithm \ref{alg:subset_search} are far superior compared to the naive front-door adjustment using $\rvbb$.

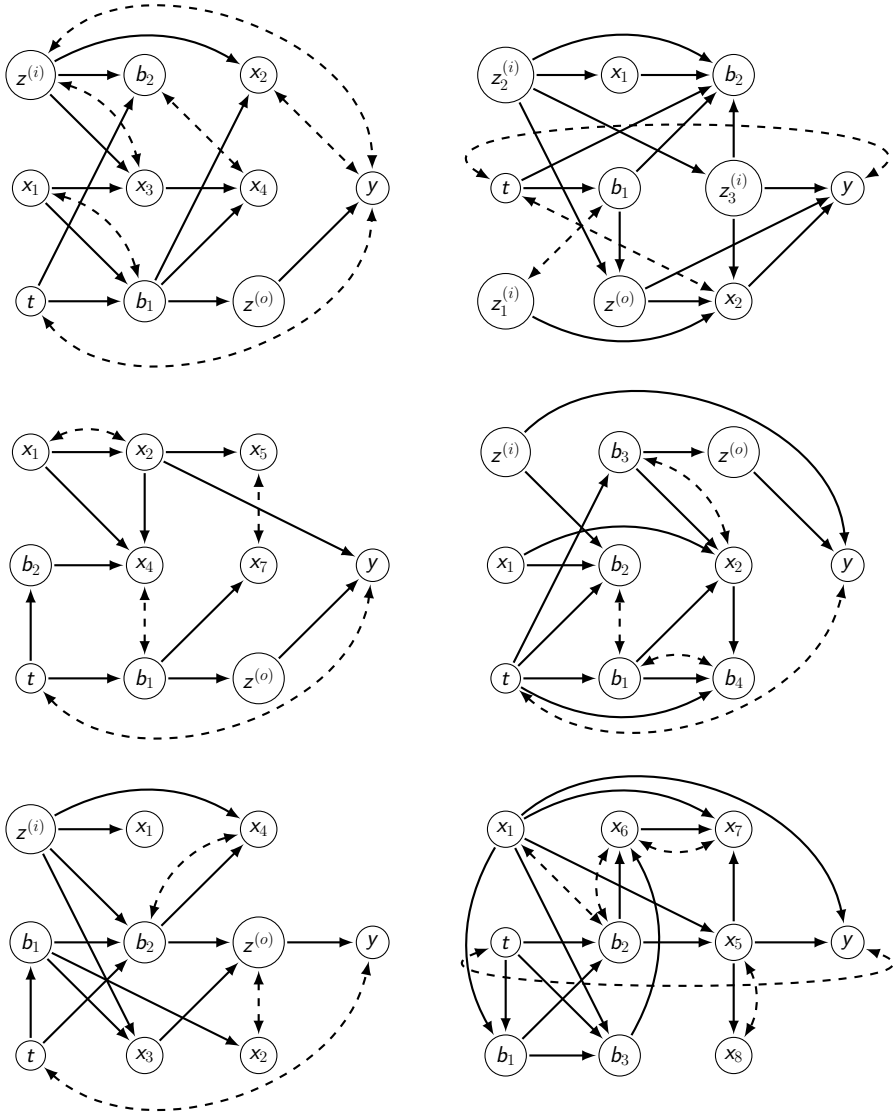
\begin{figure}[h]
  \centering
  \begin{tabular}{c}
	\begin{tikzpicture}[scale=0.5, every node/.style={transform shape}, > = latex, shorten > = 1pt, shorten < = 1pt]
	\node[shape=circle,draw=black](x1) at (0,3) {\LARGE$\rvt$};
	\node[shape=circle,draw=black](x3) at (3,3) {\LARGE$\rvb_1$};
    \node[shape=circle,draw=black](x7) at (6,3) {\LARGE$\rvz^{(o)}$};
    \node[shape=circle,draw=black](x2) at (0,6) {\LARGE$\rvx_1$};
	\node[shape=circle,draw=black](x5) at (3,6) {\LARGE$\rvx_3$};
    \node[shape=circle,draw=black](x6) at (6,6) {\LARGE$\rvx_4$};
    \node[shape=circle,draw=black](x0) at (0,9) {\LARGE$\rvz^{(i)}$};
	\node[shape=circle,draw=black](x8) at (3,9) {\LARGE$\rvb_2$};
    \node[shape=circle,draw=black](x4) at (6,9) {\LARGE$\rvx_2$};
    \node[shape=circle,draw=black](y) at (9,6) {\LARGE$\rvy$};
	\path[style=thick][->](x1) edge (x3);
    \path[style=thick][->](x3) edge (x7);
    \path[style=thick][->](x2) edge (x5);
    \path[style=thick][->](x5) edge (x6);
    \path[style=thick][->](x0) edge (x8);
    \path[style=thick][->, bend left](x0) edge (x4);
    \path[style=thick][->](x2) edge (x3);
    \path[style=thick][<->, bend left](x2) [dashed] edge (x3);
    \path[style=thick][->](x0) edge (x5);
    \path[style=thick][<->, bend left](x0) [dashed] edge (x5);
    \path[style=thick][->](x3) edge (x6);
    \path[style=thick][->](x3) edge (x4);
    \path[style=thick][<->](x6) [dashed] edge (x8);
    \path[style=thick][->](x1) edge (x8);
    \path[style=thick][->](x7) edge (y);
    \path[style=thick][<->](x4) [dashed] edge (y);
    \path[style=thick][<->, bend left=70](x0) [dashed] edge (y);
    \path[style=thick][<->, bend right=70](x1) [dashed] edge (y);
	\node[shape=circle,draw=black](g2x7) at (12.5,3) {\LARGE$\rvz^{(i)}_1$};
	\node[shape=circle,draw=black](g2x5) at (15.5,3) {\LARGE$\rvz^{(o)}$};
    \node[shape=circle,draw=black](g2x8) at (18.5,3) {\LARGE$\rvx_2$};
    \node[shape=circle,draw=black](g2x2) at (12.5,6) {\LARGE$\rvt$};
	\node[shape=circle,draw=black](g2x4) at (15.5,6) {\LARGE$\rvb_1$};
    \node[shape=circle,draw=black](g2x3) at (18.5,6) {\LARGE$\rvz^{(i)}_3$};
    \node[shape=circle,draw=black](g2x0) at (12.5,9) {\LARGE$\rvz^{(i)}_2$};
	\node[shape=circle,draw=black](g2x1) at (15.5,9) {\LARGE$\rvx_1$};
    \node[shape=circle,draw=black](g2x6) at (18.5,9) {\LARGE$\rvb_2$};
    \node[shape=circle,draw=black](g2y) at (21.5,6) {\LARGE$\rvy$};
	\path[style=thick][->](g2x0) edge (g2x1);
    \path[style=thick][->](g2x1) edge (g2x6);
    \path[style=thick][->, bend left](g2x0) edge (g2x6);
    \path[style=thick][->](g2x2) edge (g2x4);
    \path[style=thick][->](g2x2) edge (g2x6);
    \path[style=thick][->](g2x4) edge (g2x6);
    \path[style=thick][->](g2x3) edge (g2x6);
    \path[style=thick][<->](g2x2) [dashed] edge (g2x8);
    \path[style=thick][->](g2x3) edge (g2x8);
    \path[style=thick][<->, bend left=150](g2x2) [dashed] edge (g2y);
    \path[style=thick][->](g2x5) edge (g2x8);
    \path[style=thick][->, bend right](g2x7) edge (g2x8);
    \path[style=thick][<->](g2x4) [dashed] edge (g2x7);
    \path[style=thick][->](g2x0) edge (g2x3);
    \path[style=thick][->](g2x0) edge (g2x5);
    \path[style=thick][->](g2x4) edge (g2x5); 
    \path[style=thick][->](g2x5) edge (g2y);
    \path[style=thick][->](g2x8) edge (g2y);
    \path[style=thick][->](g2x3) edge (g2y);
	\node[shape=circle,draw=black](g3x3) at (0,-1) {\LARGE$\rvx_1$};
	\node[shape=circle,draw=black](g3x4) at (3,-1) {\LARGE$\rvx_2$};
    \node[shape=circle,draw=black](g3x8) at (6,-1) {\LARGE$\rvx_5$};
    \node[shape=circle,draw=black](g3x5) at (0,-4) {\LARGE$\rvb_2$};
	\node[shape=circle,draw=black](g3x6) at (3,-4) {\LARGE$\rvx_4$};
    \node[shape=circle,draw=black](g3x7) at (6,-4) {\LARGE$\rvx_7$};
    \node[shape=circle,draw=black](g3x0) at (0,-7) {\LARGE$\rvt$};
	\node[shape=circle,draw=black](g3x1) at (3,-7) {\LARGE$\rvb_1$};
    \node[shape=circle,draw=black](g3x2) at (6,-7) {\LARGE$\rvz^{(o)}$};
    \node[shape=circle,draw=black](g3y) at (9,-4) {\LARGE$\rvy$};
	\path[style=thick][->](g3x0) edge (g3x1);
    \path[style=thick][->](g3x1) edge (g3x2);
    \path[style=thick][->](g3x1) edge (g3x7);
    \path[style=thick][->](g3x5) edge (g3x6);
    \path[style=thick][->](g3x3) edge (g3x6);
    \path[style=thick][->](g3x3) edge (g3x4);
    \path[style=thick][<->, bend left](g3x3) [dashed] edge (g3x4);
    \path[style=thick][->](g3x4) edge (g3x8);
    \path[style=thick][<->](g3x7) [dashed] edge (g3x8);
    \path[style=thick][->](g3x2) edge (g3y);
    \path[style=thick][->](g3x4) edge (g3y);
    \path[style=thick][<->](g3x1) [dashed] edge (g3x6);
    \path[style=thick][->](g3x4) edge (g3x6);
    \path[style=thick][->](g3x0) edge (g3x5);
    \path[style=thick][<->, bend right=65](g3x0) [dashed] edge (g3y);
	\node[shape=circle,draw=black](g4x5) at (12.5,-1) {\LARGE$\rvz^{(i)}$};
	\node[shape=circle,draw=black](g4x2) at (15.5,-1) {\LARGE$\rvb_3$};
    \node[shape=circle,draw=black](g4x6) at (18.5,-1) {\LARGE$\rvz^{(o)}$};
    \node[shape=circle,draw=black](g4x3) at (12.5,-4) {\LARGE$\rvx_1$};
	\node[shape=circle,draw=black](g4x8) at (15.5,-4) {\LARGE$\rvb_2$};
    \node[shape=circle,draw=black](g4x4) at (18.5,-4) {\LARGE$\rvx_2$};
    \node[shape=circle,draw=black](g4x0) at (12.5,-7) {\LARGE$\rvt$};
	\node[shape=circle,draw=black](g4x1) at (15.5,-7) {\LARGE$\rvb_1$};
    \node[shape=circle,draw=black](g4x7) at (18.5,-7) {\LARGE$\rvb_4$};
    \node[shape=circle,draw=black](g4y) at (21.5,-4) {\LARGE$\rvy$};
	\path[style=thick][->](g4x0) edge (g4x1);
    \path[style=thick][->](g4x1) edge (g4x7);
    \path[style=thick][<->, bend left](g4x1) [dashed] edge (g4x7);
    \path[style=thick][->, bend right](g4x0) edge (g4x7);
    \path[style=thick][->](g4x3) edge (g4x8);
    \path[style=thick][->, bend left](g4x3) edge (g4x4);
    \path[style=thick][->](g4x2) edge (g4x6);
    \path[style=thick][<->](g4x1) [dashed] edge (g4x8);
    \path[style=thick][->](g4x1) edge (g4x4);
    \path[style=thick][->](g4x5) edge (g4x8);
    \path[style=thick][<->, bend left = 30](g4x2) [dashed] edge (g4x4);
    \path[style=thick][->](g4x0) edge (g4x2);
    \path[style=thick][->](g4x0) edge (g4x8);
    \path[style=thick][->](g4x2) edge (g4x4); 
    \path[style=thick][->](g4x6) edge (g4y);
    \path[style=thick][->, bend left=65](g4x5) edge (g4y);
    \path[style=thick][->](g4x4) edge (g4x7);
    \path[style=thick][<->, bend right=60](g4x0) [dashed] edge (g4y);
    \node[shape=circle,draw=black](g5x1) at (0,-11) {\LARGE$\rvz^{(i)}$};
    \node[shape=circle,draw=black](g5x3) at (3,-11) {\LARGE$\rvx_1$};
    \node[shape=circle,draw=black](g5x8) at (6,-11) {\LARGE$\rvx_4$};
    \node[shape=circle,draw=black](g5x2) at (0,-14) {\LARGE$\rvb_1$};
    \node[shape=circle,draw=black](g5x4) at (3,-14) {\LARGE$\rvb_2$};
    \node[shape=circle,draw=black](g5x7) at (6,-14) {\LARGE$\rvz^{(o)}$};
    \node[shape=circle,draw=black](g5x0) at (0,-17) {\LARGE$\rvt$};
    \node[shape=circle,draw=black](g5x6) at (3,-17) {\LARGE$\rvx_3$};
    \node[shape=circle,draw=black](g5x5) at (6,-17) {\LARGE$\rvx_2$};
    \node[shape=circle,draw=black](g5y) at (9,-14) {\LARGE$\rvy$};
    \path[style=thick][->](g5x2) edge (g5x4);
    \path[style=thick][->](g5x4) edge (g5x7);
    \path[style=thick][->](g5x0) edge (g5x4);
    \path[style=thick][->](g5x2) edge (g5x6);
    \path[style=thick][->](g5x1) edge (g5x6);
    \path[style=thick][->](g5x1) edge (g5x3);
    \path[style=thick][->](g5x1) edge (g5x4);
    \path[style=thick][->](g5x2) edge (g5x5);
    \path[style=thick][<->](g5x5) [dashed] edge (g5x7);
    \path[style=thick][->](g5x0) edge (g5x2);
    \path[style=thick][<->, bend left](g5x4) [dashed] edge (g5x8);
    \path[style=thick][->](g5x7) edge (g5y);
    \path[style=thick][->, bend left](g5x1) edge (g5x8);
    \path[style=thick][->](g5x6) edge (g5x7);
    \path[style=thick][->](g5x4) edge (g5x8);
    \path[style=thick][<->, bend right=60](g5x0) [dashed] edge (g5y);
    \node[shape=circle,draw=black](g6x1) at (12.5,-11) {\LARGE$\rvx_1$};
    \node[shape=circle,draw=black](g6x6) at (15.5,-11) {\LARGE$\rvx_6$};
    \node[shape=circle,draw=black](g6x7) at (18.5,-11) {\LARGE$\rvx_7$};
    \node[shape=circle,draw=black](g6x0) at (12.5,-14) {\LARGE$\rvt$};
    \node[shape=circle,draw=black](g6x3) at (15.5,-14) {\LARGE$\rvb_2$};
    \node[shape=circle,draw=black](g6x5) at (18.5,-14) {\LARGE$\rvx_5$};
    \node[shape=circle,draw=black](g6x2) at (12.5,-17) {\LARGE$\rvb_1$};
    \node[shape=circle,draw=black](g6x4) at (15.5,-17) {\LARGE$\rvb_3$};
    \node[shape=circle,draw=black](g6x8) at (18.5,-17) {\LARGE$\rvx_8$};
    \node[shape=circle,draw=black](g6y) at (21.5,-14) {\LARGE$\rvy$};
    \path[style=thick][->](g6x2) edge (g6x4);
    \path[style=thick][->](g6x2) edge (g6x3);
    \path[style=thick][->](g6x1) edge (g6x4);
    \path[style=thick][->](g6x1) edge (g6x5);
    \path[style=thick][<->, bend left](g6x5) [dashed] edge (g6x8);
    \path[style=thick][->, bend left](g6x1) edge (g6x7);
    \path[style=thick][->, bend right](g6x1) edge (g6x2);
    \path[style=thick][->](g6x3) edge (g6x6);
    \path[style=thick][->, bend right](g6x4)  edge (g6x6);
    \path[style=thick][->](g6x0) edge (g6x2);
    \path[style=thick][<->, bend left](g6x3) [dashed] edge (g6x6);
    \path[style=thick][<->, bend right](g6x6) [dashed] edge (g6x7);
    \path[style=thick][->](g6x0) edge (g6x3);
    \path[style=thick][->](g6x0) edge (g6x4);
    \path[style=thick][<->](g6x1) [dashed] edge (g6x3);
    \path[style=thick][->](g6x3) edge (g6x5);
    \path[style=thick][->](g6x5) edge (g6y);
    \path[style=thick][->](g6x5) edge (g6x7); 
    \path[style=thick][->](g6x5) edge (g6x8);
    \path[style=thick][->, bend left=60](g6x1) edge (g6y);
    \path[style=thick][->](g6x6) edge (g6x7);
    \path[style=thick][<->, bend right=160](g6x0) [dashed] edge (g6y);
	\end{tikzpicture}
 \end{tabular}
 \caption{The SMCMs used in Section \ref{sec:descsynth} to compare Algorithm \ref{alg:subset_search} with the \texttt{Baseline} that uses $\rvbb$ for front-door adjustment. These are the 6 out of the 100 random graphs in Section \ref{subsec_random_graphs} for $p = 10$, $d = 2$, and $q = 1.0$ where our approach was successful indicating existence of $\rvbz = (\rvbzi,\rvbzo)$ such that the conditional independence statements in Theorem \ref{thm_main} hold.}
 \label{fig_random_smcms}
 \end{figure}

In Figure \ref{fig_random_smcms}, we provide the 6 random SMCMs used in Section \ref{sec:descsynth}. As mentioned in Section \ref{subsec_random_graphs}, we choose the last variable in the causal ordering as $\rvy$ and a variable that is ancestor of $\rvy$ but not its parent or grandparent as $\rvt$. We also show the corresponding $\rvbz = (\rvbzi, \rvbzo)$ satisfying \cref{eq_ci,eq_condition}.

\subsection{German Credit dataset}
As in Section \ref{subsec_expts_real_world}, we assess the conditional independence associated with the selected $\rvbz$ for the choice of $\rvbb\!=\! \{$\# of people financially dependent on the applicant, applicant's savings$\}$, Algorithm \ref{alg:subset_search} results in $\rvbzi\!=\! \{$purpose for which the credit was needed, applicant’s checking account status with the bank$\}$ via 100 random bootstraps. We show the corresponding p-values for these bootstraps in a histogram in Figure \ref{fig_german_2} below. As expected, we observe the p-values to be spread out. 

\begin{figure}[h]
    \centering
    \begin{tabular}{c}
    \includegraphics[width=0.75\linewidth,clip]{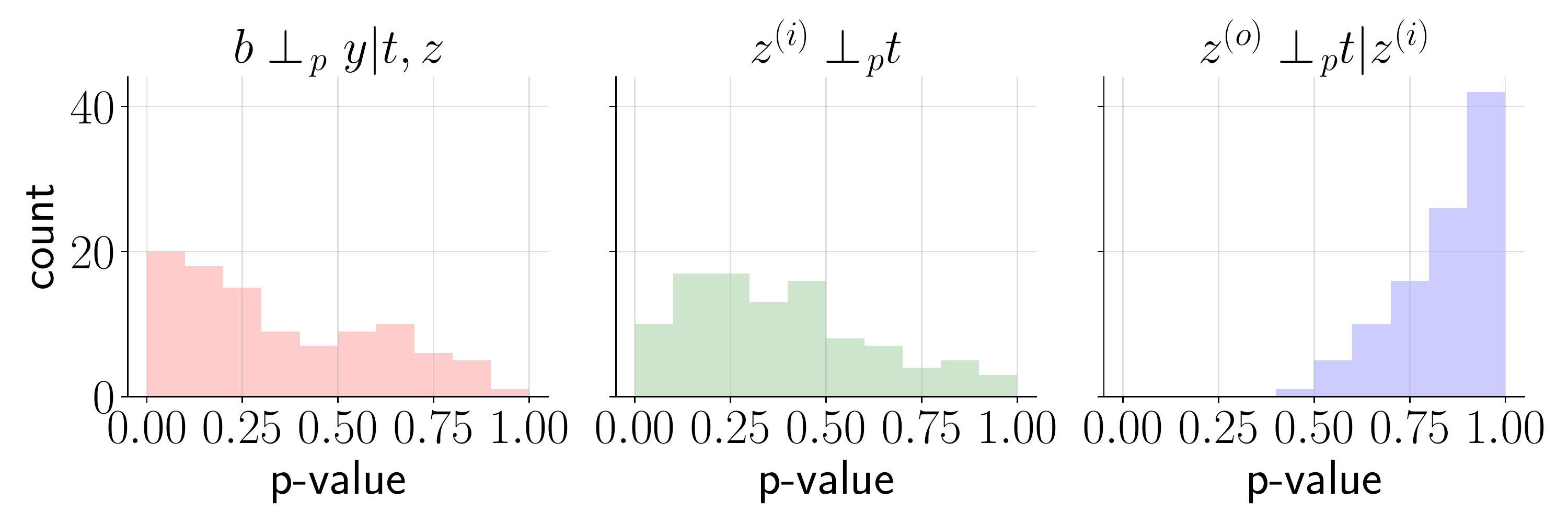}
    \end{tabular}
\caption{Histograms of p-values of the conditional independencies in \eqref{eq_thm_main_1} and \eqref{eq_thm_main_2} over 100 bootstrap runs for $\rvbb\!=\! \{$\# of people financially dependent on the applicant, applicant's savings$\}$, Algorithm \ref{alg:subset_search} results in $\rvbzi\!=\! \{$purpose for which the credit was needed, applicant’s checking account status with the bank$\}$.}
\label{fig_german_2}
\end{figure}

\subsection{Adult dataset}
The Adult dataset \citep{adult1996} is used for income analysis where the goal is to predict whether an individual’s income is more than \$50,000 using 14 demographic and socio-economic features. The sensitive attribute $\rvt$ is the individual's sex, either male or female. Further, the categorical attributes are one-hot encoded. As with German Credit dataset, we apply Algorithm \ref{alg:subset_search} with $n_r = 100$ and $p_v = 0.1$ where we search for a set $\rvbz = (\rvbzo, \rvbzi)$ of size at most $3$ under the following two assumptions on the set of all children $\rvbb$ of $\rvt$: (1) $\rvbb = \{$\# individual's relationship status (which includes wife/husband)$\}$ and (2) $\rvbb = \{$\# individual's relationship status (which includes wife/husband), individual's occupation$\}$. In either case, Algorithm \ref{alg:subset_search} was unable to find a suitable $\rvbz$ satisfying $\rvbb \indep \rvy | \rvbz, \rvt$. This suggests that in this dataset, there may not be any non-child descendants of the sensitive attribute, which is required for our criterion to hold.

\subsection{Licenses}
In this work, we used a workstation with an AMD Ryzen Threadripper 3990X 64-Core Processor (128 threads in total) with 256 GB RAM and 2x Nvidia RTX 3090 GPUs. However, our simulations only used the CPU resources of the workstation.

We mainly relied on the following Python repositories --- (a) networkx (\url{https://networkx.org}), (b) causal-learn (\url{https://causal-learn.readthedocs.io/en/latest/}), (c) RCoT \citep{strobl2019approximate} and (d) ridgeCV, (\url{https://github.com/scikit-learn/scikit-learn/tree/15a949460/sklearn/linear_model/_ridge.py}). We did not modify any of the code under licenses; we only installed these repositories as packages. 

In addition to these, we used two public datasets  (a) German Credit dataset (\url{https://archive.ics.uci.edu/ml/datasets/statlog+(german+credit+data)}) and (b) Adult dataset (\url{https://archive.ics.uci.edu/ml/datasets/adult}). These datasets are commonly used benchmark datasets for causal fairness, which is why we chose them for our comparisons.

\end{document}